\PassOptionsToPackage{pagebackref,breaklinks,colorlinks}{hyperref}

\documentclass[10pt,twocolumn,letterpaper]{article}
\usepackage[accsupp]{axessibility}
\usepackage{wacv}              

\usepackage{graphicx}
\usepackage{amsmath}
\usepackage{amssymb}
\usepackage{booktabs}

\usepackage{orcidlink}
\usepackage{amsmath}

\usepackage{subcaption}
\usepackage{multirow}
\usepackage{color}
\usepackage{algorithm}
\usepackage{algpseudocode}
\usepackage{float}
\newtheorem{definition}{Definition}

\usepackage{hyperref}

\usepackage[capitalize]{cleveref}
\crefname{section}{Sec.}{Secs.}
\Crefname{section}{Section}{Sections}
\Crefname{table}{Table}{Tables}
\crefname{table}{Tab.}{Tabs.}

\def\wacvPaperID{2696} 
\def\confName{WACV}
\def\confYear{2025}

\begin{document}

\title{On Explaining Knowledge Distillation: Measuring and Visualising the Knowledge Transfer Process}

\author{Gereziher Adhane\\
Universitat Oberta de Catalunya\\
{\tt\small gradhane@uoc.edu}
\and
Mohammad Mahdi Dehshibi\\
Universidad Carlos III de Madrid\\
{\tt\small mohammad.dehshibi@yahoo.com}
\and
Dennis Vetter\\
Goethe University Frankfurt\\
{\tt\small vetter@em.uni-frankfurt.de}
\and
David Masip\\
Universitat Oberta de Catalunya\\
{\tt\small dmasipr@uoc.edu}
\and
Gemma Roig\\
Goethe University Frankfurt\\
{\tt\small roig@cs.uni-frankfurt.de}
}

\maketitle

\begin{abstract} 

Knowledge distillation (KD) remains challenging due to the opaque nature of the knowledge transfer process from a Teacher to a Student, making it difficult to address certain issues related to KD. To address this, we proposed \textit{UniCAM}, a novel gradient-based visual explanation method, which effectively interprets the knowledge learned during KD. Our experimental results demonstrate that with the guidance of the Teacher's knowledge, the Student model becomes more efficient, learning more relevant features while discarding those that are not relevant. We refer to the features learned with the Teacher's guidance as distilled features and the features irrelevant to the task and ignored by the Student as residual features. Distilled features focus on key aspects of the input, such as textures and parts of objects. In contrast, residual features demonstrate more diffused attention, often targeting irrelevant areas, including the backgrounds of the target objects. In addition, we proposed two novel metrics: the \textit{feature similarity score (FSS)} and the \textit{relevance score (RS)}, which quantify the relevance of the distilled knowledge. Experiments on the CIFAR10, ASIRRA, and Plant Disease datasets demonstrate that \textit{UniCAM} and the two metrics offer valuable insights to explain the KD process.

\end{abstract}
\section{Introduction}
Knowledge Distillation (KD) has emerged as a crucial technique in deep learning, especially in computer vision. It aims to develop efficient models without compromising performance~\cite{Gou2023Multilevel,chung2020feature,zhang2018deep,passban2021alp,cho2019efficacy}. By transferring knowledge from typically a complex Teacher model to a simpler Student model, KD can potentially address the increasing demand for deploying robust yet lightweight models in practical scenarios~\cite{zagoruyko2017paying}. Nevertheless, despite its widespread adoption, the underlying mechanisms of KD remain somewhat opaque, thus impeding its broader application and theoretical comprehension.

The current research in knowledge distillation (KD) is confronted with four main challenges, including (1) understanding the specific knowledge that is transferred from Teacher to Student~\cite{cheng2020explaining}; (2) evaluating whether KD improves the Student's focus on task-relevant features compared to independent training~\cite{wang2019deepvid,chung2020feature}; (3) measuring the importance of features adopted or ignored by the Student for the target task; (4) addressing and resolving KD failures, mainly when there are significant architectural differences between Teacher and Student models~\cite{Son2021ICCV,stanton2021does,mirzadeh2020improved}.

Existing visual explainability methods for Convolutional Neural Networks (CNNs), like Grad-CAM~\cite{selvaraju2017grad}, are not equipped to tackle these KD-specific challenges. While effective for single-model predictions, these methods cannot capture the nuance of knowledge transfer between models or quantify the relevance of distilled knowledge. Specifically, Grad-CAM focuses on the importance of class-specific features within a single model. However, it does not distinguish between knowledge inherited from the Teacher and knowledge independently learned by the Student.

To address these issues, we introduce a new framework for improving the explainability of KD. We will first define the key terms used in this paper to ensure clarity. \textbf{Distilled features} are unique to the Student and are acquired through KD-based training, which the Student considers relevant to the task. \textbf{Residual features} are present in the Teacher (Base model) but are not adopted by the Student, as the Student finds them irrelevant to the task during KD-based training. We used \textbf{Unique features} to collectively refer to the distilled and residual features, as they are unique to the Student and the Base model, respectively. Throughout this text, we will use the term \textbf{Base model} to refer to the model which has the same architecture as the Student but is trained using only raw data. When the Teacher and Student have similar architecture, then the Teacher will act as a suitable Base model.

Our framework consists of two main components:
\begin{itemize}
    \item Visual explainability tool – We introduce UniCAM (Unique Class Activation Mapping), a gradient-based visual explanation method designed explicitly for KD scenarios. UniCAM distinguishes between distilled and residual features using partial distance correlation to isolate unique features learned by each model. (detailed methodology in Section~\ref{sec:UniCam}). 
    \item Knowledge transfer metrics – We introduce Feature Similarity Score (FSS) and Relevance Score (RS). The former, FSS, quantifies the alignment of attention patterns between Student and Teacher (Base model) using distance correlation. The latter, RS, evaluates the task relevance of distilled and residual features using distance correlation between the extracted features and BERT embeddings of the ground truth labels~\cite{devlin2018bert}.
\end{itemize}

FSS and RS work in tandem to provide a comprehensive analysis of KD's effectiveness. While FSS measures the similarity of learned representations between the Base model and the Student, RS assesses their relevance to the target task. This dual approach is crucial because high similarity (FSS) does not always guarantee optimal task-specific learning (RS). For example, a Student might closely imitate a Teacher's attention patterns (high FSS) but struggle to differentiate between important and irrelevant features for the specific task (potentially low RS).

Our key contributions are fourfold:
\begin{itemize}
    \item Introducing \textbf{Uni}que \textbf{C}lass \textbf{A}ctivation \textbf{M}aps (\textit{UniCAM}), a novel method for visualising the knowledge transfer process in KD. This approach offers insights into how Students acquire relevant features and discard less important ones under the Teacher's guidance, addressing challenges 1 and 2.
    \item Proposing the complementary FSS and RS metrics, which enable quantitative analysis of KD effectiveness. This dual-metric approach captures both the similarity of learned representations (FSS) and their task-specific relevance (RS), addressing challenges 2 and 3.
    \item Conducting extensive experiments across various KD techniques and model architectures, demonstrating the broad applicability of our approach. Our evaluation encompasses standard image classification datasets (CIFAR-10~\cite{krizhevsky2009learning}, Microsoft PetImages~\cite{elson2007asirra}) and a more challenging fine-grained plant disease classification task~\cite{HughesS15}. This comprehensive assessment validates the effectiveness of our methods across diverse scenarios.
    \item Providing a detailed examination of KD failure cases, particularly those arising from significant capacity gaps between Teacher and Student models. Through this analysis, we demonstrate how our method can guide the selection of appropriate Teacher models or intermediate architectures (Teacher assistants) to improve KD outcomes, addressing challenge 4.
\end{itemize}


 \section{Related Work}
\label{sec:Relatedworks}
Recent works on KD have focused on improving the performance of the Student~\cite{jiao2020tinybert}, adapting the distillation process to specific tasks~\cite{douillard2020podnet}, or developing alternative methods for knowledge transfer~\cite{heo2019overhaul}. In contrast to the performance-oriented works, some studies have explored KD explainability using various techniques. For instance, Cheng et al.~\cite{cheng2020explaining} used information theory and mutual information to visualise and measure knowledge during KD. However, this method requires human annotations of the background and foreground objects, which limits its applicability and scalability. Moreover, using entropy to quantify randomness might be unreliable in scenarios with highly correlated data or multiple modes. Similarly, Wang et al.~\cite{Xue2021KDExplainerAT} used the KD and generative models to diagnose and interpret image classifiers, but this approach does not account for the knowledge acquired by the Student.

Existing visual explainability techniques offer valuable insights into how CNNs make decisions (e.g.,~\cite{arrieta2020explainable,dehshibi2023advise}). Applying these methods to KD could reveal if the Student focuses on the same input areas as a Base model and learns similar or superior features. For example, DeepVID~\cite{wang2019deepvid} visually interprets and diagnoses image classifiers through KD. Haselhoff et al.~\cite{haselhoff2021towards} proposed a probability density encoder and a Gaussian discriminant decoder to describe how explainers deviate from concepts' training data in KD. However, existing visual explainability techniques cannot visualise the saliency maps of distilled and residual features. Similarity metrics offer a promising strategy to measure and identify features unique to one model~\cite{zhen2022versatile}, which could be useful to effectively quantify and explain the distilled knowledge.

Similarity measures have been widely employed across various disciplines, including machine learning~\cite{dehshibi2021deep,dehshibi2023pain,adhane2022incorporating,dehshibi2024beenet}, information theory~\cite{dehshibi2021electrical,dehshibi2021stimulating,gholami2020novel}, and computational neuroscience~\cite{kriegeskorte2018cognitive}. These measures offer valuable insights into how information is processed and encoded in different contexts. In deep learning, similarity metrics have been useful to (1) quantify how DNNs replicate the brain's encoding process~\cite{wen2018neural}, (2) compare vision transformers and convnets~\cite{raghu2021vision,kornblith2019similarity}, (3) gain insights into transfer learning~\cite{dwivedi2019representation,neyshabur2020being}, and (4) explain the mechanisms behind deep model training~\cite{gotmare2019closer}. In this study, we propose a novel gradient-based visual explainability technique and quantitative metrics that leverage similarity measures to isolate unique features of each model and quantify their relevance, enhancing the transparency of the KD process.

\section{Methodology}
\label{sec:methodology}
Given a Student model trained using KD and a Base model trained solely on data, our objective is to explain and quantify the amount of knowledge distilled during KD. Our approach leverages gradient-based explainability techniques to compute gradients with respect to input features, along with distance correlation (dCor)\cite{szekely2014partial} and partial distance correlation (pdCor)\cite{szekely2014partial,zhen2022versatile}. dCor measures the dependence between two random vectors that capture their multidimensional associations. Similarly, pdCor extends dCor to measure the association between two random vectors after adjusting for the influence of a third vector. It is computed by projecting the distance matrices onto a Hilbert space and taking the inner product between the U-centered matrices. Zhen et al.~\cite{zhen2022versatile} used pdCor to condition multiple models and identify their unique features\footnote{Unique features are the features specific either to the Base model (residual features) or to the Student (distilled features).}, which means removing the common features and assessing the remaining ones. This enables us to introduce a novel visual explanation and metrics to assess the knowledge the student learned (distilled features) and that it may have overlooked (residual features).

\subsection{UniCAM: {Uni}que {C}lass {A}ctivation {M}apping}
\label{sec:UniCam}
Our goal is to generate saliency maps that highlight the \emph{distilled} and \emph{residual} features of the Student model, emphasising their importance and revealing their attention patterns to enable a deeper understanding of KD. Existing gradient-based visual explanations~\cite{selvaraju2017grad,arrieta2020explainable,dehshibi2023advise} generates saliency maps based on the gradients of the target class, effectively revealing the relevance of features for the target prediction. However, these techniques are not suited for KD, as they do not identify the distilled or residual features compared to the Base model. To overcome this limitation, we introduce \textit{UniCAM}, a novel gradient-based explainability technique tailored for KD. \textit{UniCAM} leverages pdCor to adjust feature representations and remove the shared features between the Student and the Base model. This process isolates the \emph{distilled} and \emph{residual} features, which correspond to the knowledge the Student has acquired or overlooked, providing insights into the relevance of these features for the target task.

Let $x_s$ and $x_b$ represent the features extracted from a specific convolutional layer of the Student and the Base model, respectively. The \textit{UniCAM} method follows the following key steps: (1) First, we compute the pairwise distance matrices for both $x_s$ and $x_b$, which capture the relationships between different feature vectors within the Student and Base models. (2) Next, we normalise these distance matrices to create adjusted distance matrices, denoted as $P^{s}$ and $P^{b}$, which ensure the distance information is centred and standardised. (3) We then calculate the mutual influence between the Student and Base models' features and remove the shared components, effectively isolating the unique features that each model has learned. (4) Finally, we generate heatmaps of these unique features, which localise the importance and relevance of the distilled features compared to the Base model.

Following the approach explained in \cite{szekely2014partial}, we first compute the pairwise distance matrix $D^{(s)} = (D^{(s)}_{i,j})$ for the Student's feature set. The pairwise distance matrix captures the relationship between every pair of feature vectors within the Student model:
\begin{equation}
    D^{(s)}_{i,j} = \sqrt{(x_i - x_j)^2 + \epsilon},
\end{equation}
where $\epsilon$ is a small  positive number added for numerical stability. This matrix helps us quantify how closely related the feature vectors are, which forms the basis for identifying unique and shared features.

Next, we normalise the distance matrix using Eq.~\ref{eq:u-center} to obtain the adjusted distance matrix $P^{(s)}$. This normalisation is a U-centred projection, which centres the matrix around the mean and adjusts it for the overall distribution of distances.
\small
\begin{equation}
    P^{(s)}_{i,j} = \left\{\begin{aligned}
        D^{(s)}_{i,j} &- \frac{1}{n-2} \sum_{l=1}^n D^{(s)}_{i, l} - \frac{1}{n-2} \sum_{k=1}^n D^{(s)}_{k,j} &\\
        &+ \frac{1}{(n-1)(n-2)} \sum_{k=1}^n \sum_{l=1}^n D^{(s)}_{k,l}, & i \neq j; \\
        0,& & i = j.
        \end{aligned}\right.
\label{eq:u-center}
\end{equation}
\normalsize

This step ensures that the distance information is properly centred and scaled, making it easier to compare features across models. To isolate the unique features learned by the Student model, we adjust for the mutual influence between the Student and Base models. This step subtracts the shared features between the Student and Base model:
\begin{equation}
    x_{s|unique} = P^{(s)} - \frac{\langle P^{(s)}, P^{(b)} \rangle}{\langle P^{(b)}, P^{(b)} \rangle} \cdot P^{(b)}.
\end{equation}

Here, we compute the inner product between the adjusted distance matrices of the Student and Base models, $\langle P^{(s)}, P^{(t)} \rangle$, which captures their shared information as:
\small
\begin{equation}
    \langle P^{(s)} , P^{(b)}  \rangle = \frac{1}{n(n-3)} \sum_{i \neq j}\left( P^{(s)}_{i,j} \cdot P^{(b)} _{i,j} \right).
\end{equation}
\normalsize
We subtract the common features to isolate the unique ones in each model (distilled and residual features) and reflect what it has learned beyond the Base model.

Once the unique features are extracted, we compute their importance for the target task prediction by calculating the gradients of the prediction with respect to these features. The gradient-based importance of each unit $k$ for class $c$ is given by:
\begin{equation}
\beta_k^{(x_{s|unique}, c)} = \frac{1}{N}\sum_i \sum_j \frac{\partial y^c}{\partial A_{ij}^{(x_{s|unique})}},
\label{eq:weightunit}
\end{equation}
where $A_{ij}$ represents the activation at position $(i,j)$ in the feature map, $y^c$ is the output score for class $c$, $\beta_k^{(x_{s|unique}, c)}$ is the weight of unit $k$ for class $c$ calculated from the unique features $x_{s|unique}$, and $N$ is a normalisation factor.

Finally, we generate the \textit{UniCAM} saliency map by combining these weights and applying a ReLU function to highlight the most important areas of the input image:
\small
\begin{equation}
    L_{\text{\textit{UniCAM}}}^{(x_{s|unique}, c)} = \text{ReLU} \left( \sum_k \beta_k^{(x_{s|unique}, c)} A^{(x_{s|unique})} \right).
\label{eq:CAMpdCor}
\end{equation}
\normalsize
This process creates a heatmap that visualises the important features the Student learned from the Teacher, making the KD process more explainable. To identify the residual features, we follow the same procedure, but instead of the Student, we consider the Base model in the previous steps. These residual features represent areas where the Student, with the guidance of the Teacher's knowledge, has determined that certain features (such as background elements or irrelevant parts of the object) are not important for the target task. Analysing residual features is crucial to understanding whether the Student is effectively ignoring irrelevant features or potentially overfitting. The above procedure is summarised in the algorithm~\ref{alg:unicam}.

\begin{algorithm}[t!]
\scriptsize
\caption{UniCAM: Unique Class Activation Mapping}
\label{alg:unicam}
\begin{algorithmic}[1]
\Require
\begin{itemize}
\item[] $x_s, x_b$ -- Features from Student and Base models.
\end{itemize}
\Ensure
\begin{itemize}
\item[] \textit{UniCAM} maps for distilled and residual features.
\end{itemize}
\State Compute pairwise distance matrix for $x_s$:
\For{$i,j = 1$ to $n$}
\State $D^{(s)}_{i,j} = \sqrt{(x_{s_i} - x_{s_j})^2 + \epsilon}$
\EndFor
\State Normalise the distance matrix to compute $P^{(s)}$ using Eq.~\ref{eq:u-center}
\State Extract the distilled features by adjusting for the mutual influence:
\State $x_{s|unique} = P^{(s)} - \frac{\langle P^{(s)}, P^{(b)} \rangle}{\langle P^{(b)}, P^{(b)} \rangle} \cdot P^{(b)}$
\State Compute the importance of the distilled features:
\State $\beta_k^{(x_{s|unique}, c)} = \frac{1}{N} \sum_i \sum_j \frac{\partial y^c}{\partial A_{ij}^{(x_{s|unique})}}$
\State Generate \textit{UniCAM}:
\State $L_{\text{\textit{UniCAM}}}^{(x_{s|unique}, c)} = \text{ReLU} \left( \sum_k \beta_k^{(x_{s|unique}, c)} A^{(x_{s|unique})} \right)$ \\
\Return \textit{UniCAM} maps for $x_{s|unique}$
\end{algorithmic}
\end{algorithm}

\subsection{Quantitative analysis of KD features}
While visualising the heatmaps using \textit{UniCAM} provides insights to make the KD process transparent, it is equally important to quantify the relevance and significance of the distilled and residual features. To this end, we introduce two novel metrics: Feature Similarity Score (\textit{FSS}) and Relevance Score (\textit{RS}). These metrics allow us to evaluate both the overall features learned by the Student compared to the Base model, as well as the distilled and residual features.

\begin{figure}[!ht]
    \centering
    \includegraphics[width=0.95\linewidth]{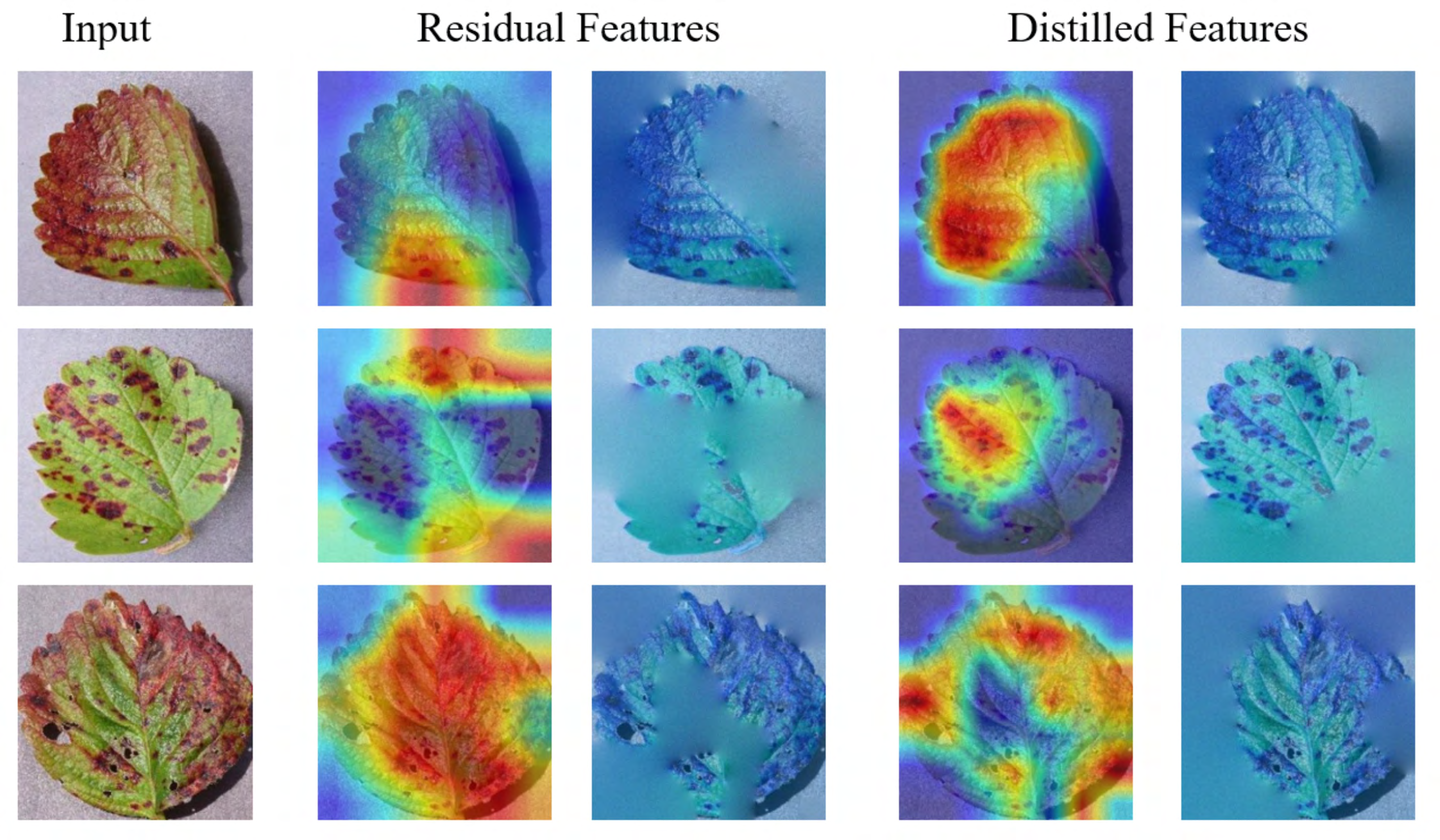}
    \caption{Residual and distilled features after perturbation.}
    \vspace{-0.5em}
    \label{Fig:featext}
\end{figure}

 To compute these metrics, we first extract the relevant features from the salient regions identified by \textit{UniCAM}. Next, we apply a perturbation technique proposed by Rong et al.\cite{rong2022consistent}, which modifies image pixels based on their prediction relevance. This perturbation preserves the most important pixels and replaces the irrelevant ones with the weighted average of their neighbours. As a result, the perturbed images retain the most salient features while reducing noise and redundancy. Fig.~\ref{Fig:featext} shows this process with examples of input images, \textit{UniCAM} explanations, and perturbed images.

The feature extraction function takes the perturbed images in Fig.~\ref{Fig:featext} as input and extract the features (See Eq.~\ref{eq:ft}) from the corresponding layer as:
\begin{equation} 
    \label{eq:ft} 
    \begin{split} 
        \hat{x}_s = f_s(I \odot \mathcal{H}),\ 
        \hat{x}_b = f_b(I \odot \mathcal{H}), 
    \end{split}
\end{equation}
where $f_s$ and $f_b$ are the feature extractor functions for the Student and Base model, respectively, $I$ is the input image, $\mathcal{H}$ is the heatmap generated by \textit{UniCAM}, $\odot$ is the element-wise multiplication operator. These features are the numerical representations of the perturbed image that capture the essential information for prediction, and their dimension depends on the number of filters and the size of the activation maps in each layer. Hence, using these features, we quantify the similarity of the attention patterns and the relevance of distilled and residual features.

\subsubsection{Feature similarity score (FSS)} 
\textit{FSS} is designed to quantify the degree of alignment between the features learned between the Student and Base model at a specific layer. Since the Student is trained with the guidance of the Teacher's knowledge, \textit{FSS} provides insight into how much the Student's focus has shifted or aligned with the Base model features. A higher \textit{FSS} value suggests that the Student and Base models are focusing on similar regions of the input, indicating that the Teacher’s knowledge has not significantly altered the core feature focus of the Student or that the task is such that both models naturally converge on similar important features. Conversely, a lower \textit{FSS} would suggest that the Student has diverged, potentially learning a more refined or generalised feature representation due to the knowledge from the Teacher. The \textit{FSS} is computed as follows:
\small
\begin{equation}
    \label{eq:fss}
    \begin{split}
    FSS  = R^2(\hat{x}_s, \hat{x}_b) = \frac{1}{k}\sum_{i=1}^{k} \text{dCor}(\hat{x}_{s_i}, \hat{x}_{b_i}),
    \end{split}
\end{equation}
\normalsize
where $k$ is the number of batches, $\hat{x}_{s_i}$ and $\hat{x}_{b_i}$ are the mini-batch features of the Student and Base model. \textit{FSS} ranges from 0 to 1, where 0 means no similarity and values close to 1 indicate higher attention pattern similarity. 

\subsubsection{Relevance score (RS)}
While \textit{FSS} measures the similarity between the features learned by the Student and Base model, it does not quantify how relevant these features are to the target task. To address this, we propose the Relevance Score (\textit{RS}), which evaluates the relevance of the distilled and residual features for the target task. 

To capture the semantic information of the target task more effectively, we use a pre-trained \textit{BERT} embedding of the ground truth labels~\cite{devlin2018bert}. Unlike traditional one-hot encodings, which provide limited information, \textit{BERT} embedding represents the labels in a high-dimensional space that captures richer semantic relationships between different labels. This allows us to compute meaningful correlations between the feature vectors encoded by the models and the ground truth, offering a more robust measure of relevance for the task. Hence, we compute the \textit{RS} as follows:
\small
\begin{equation}
    \label{eq:rs} 
    \begin{split}
        RS = R^2(\hat{x}_s, gt) = \frac{1}{k}\sum_{i=1}^{k} \text{dCor}(\hat{x}_{s_i}, {gt}_{i}),
    \end{split}
\end{equation}
\normalsize
where $\hat{x}_{s_i}$ is the features extracted by the Student for each minibatch, and ${gt}_i$ is the ground truth \textit{BERT} embeddings for the corresponding targets in each batch. To compute the \textit{RS} for the Base model, we replace $\hat{x}_{s_i}$ with $\hat{x}_{b_i}$.

Both \textit{FSS} and \textit{RS} provide a comprehensive quantitative technique to evaluate the similarity of the attention patterns and the relevance of the features learned during KD. This helps us understand whether the Student is acquiring features that are both similar and meaningful for the target task, offering deeper insights into explaining the KD process.

\section{Experiments}
\subsection{Datasets and Implementation Details}
We evaluate the proposed method on three public datasets for image classification: ASIRRA (Microsoft PetImages)~\cite{elson2007asirra}, CIFAR10~\cite{krizhevsky2009learning} and Plant disease classification dataset~\cite{HughesS15}. ASIRRA contains 25,000 images of cats and dogs, while CIFAR10 contains 60,000 images of 10 classes. These datasets are widely used as benchmarks for image classification tasks and have different levels of complexity and diversity. Plant disease classification has a more challenging and realistic problem than fine-grained image classification, where the differences between classes are subtle and require more attention to detail. More results of plant disease classification are provided in the supplementary material \textbf{Sec. A}. 

We performed various experiments to analyse and explain the KD process. First, we used ResNet-50~\cite{he2016deep} as both the Student and Teacher models, effectively making the Teacher a Base model. This allows us to isolate the effects of KD without introducing the complexity bias that can arise when using a more powerful Teacher model. It ensures that any observed differences are due to the KD process itself rather than architectural disparities between the Teacher and Student. This experiment addresses key questions (1)-(3) by analysing the performance of the Student and the Base model, the similarity in attention patterns, and the relevance of the distilled and residual features. In the second experiment, we analysed different combinations of ResNet-18, ResNet-50, and ResNet-101 as Teacher and Student models to address the \textit{fourth} question, which explores the impact of architecture differences on KD. We applied our approach to three state-of-the-art KD methods for classification: response-based KD~\cite{hinton2015distilling}, overhaul feature-based KD~\cite{heo2019overhaul}, and attention-based KD~\cite{passban2021alp}. We implemented\footnote{Code is available \href{https://github.com/gadhane/UniCAM}{https://github.com/gadhane/UniCAM}} the proposed method using PyTorch~\cite{paszke2017automatic} and open source libraries from KD~\cite{shah2020kdlib}, pdCor~\cite{zhen2022versatile} and Grad-CAM~\cite{jacobgilpytorchcam}.

\subsection{Results}
\begin{figure}[!ht]
    \centering 
    \begin{subfigure}{\linewidth} 
    \includegraphics[width=\linewidth]{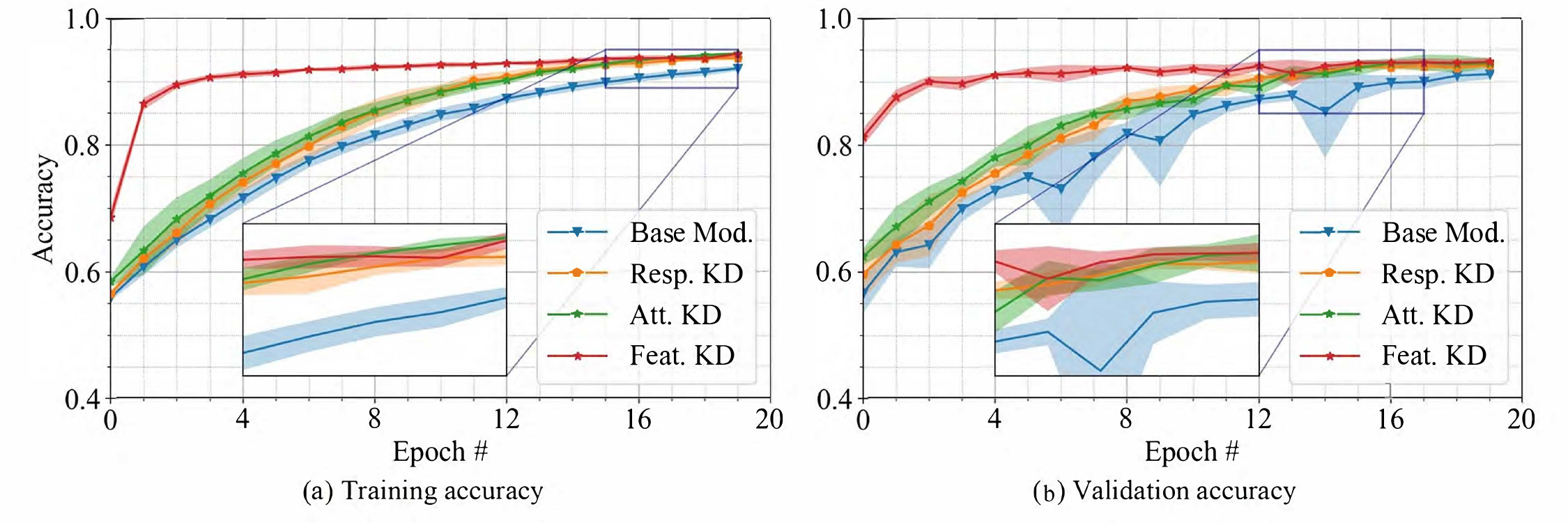} 
    \end{subfigure} 
    \begin{subfigure}{\linewidth} 
    \includegraphics[width=\linewidth]{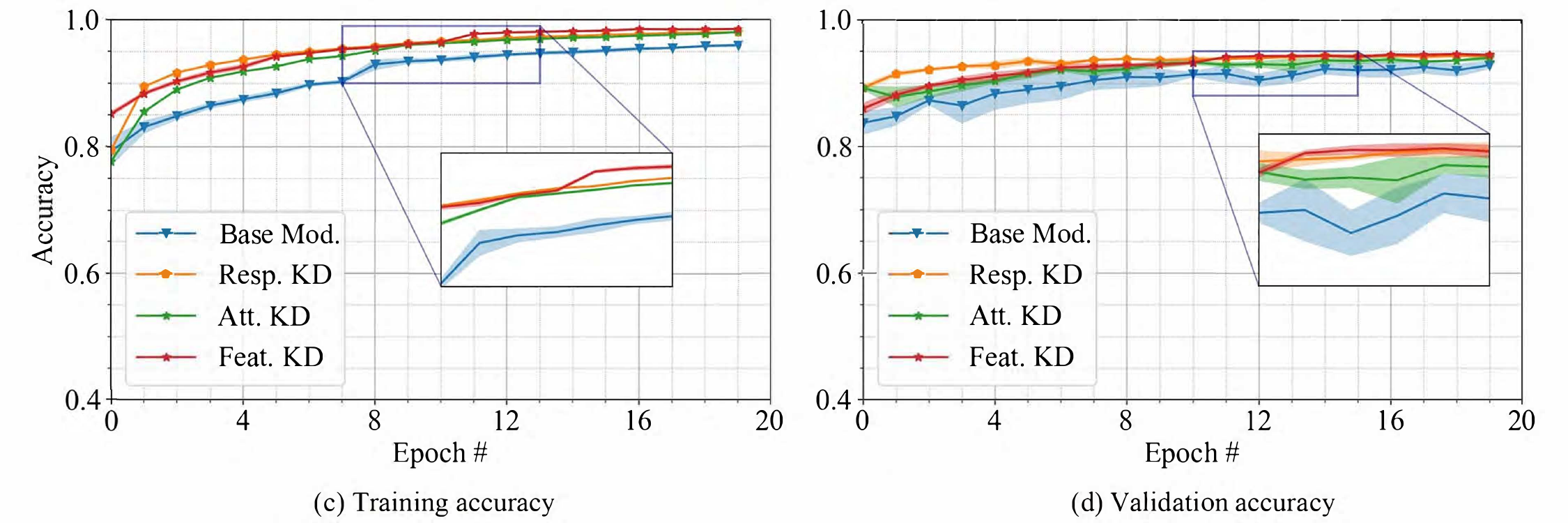} 
    \end{subfigure}
    \caption{The training and validation accuracy( (a) and (b) ASIRRA, (c) and (d) CIFAR10. The shaded region is the standard deviation.}
    \vspace{-0.5em}
    \label{Fig:trainperf} 
\end{figure}

We trained the models using 5-fold cross-validation. Details about each training setting are given in the supplementary material \textbf{Sec. D}. We assessed the performance and visual explanations of the Student and Base model trained with different KD. As shown in Fig.~\ref{Fig:trainperf}, the models trained with KD achieved higher accuracy compared to the equivalent Base model.

\subsubsection{Comparison of Student and Base model attention patterns}
We hypothesise that KD enhances the Student model’s ability to learn more relevant features while discarding irrelevant ones. To test this hypothesis, we begin by using Grad-CAM, a widely adopted explainability technique, to provide an initial qualitative comparison of the feature localisation in both the Student and Base model across different layers. While Grad-CAM alone does not explain the KD process, it serves as a first step to visually compare which model encodes more relevant features based on attention maps. This initial visualisation offers insight into the general behaviour of the models, which is then complemented by our proposed quantitative analysis using \textit{FSS} and \textit{RS}. These metrics allow us to precisely measure the similarity and relevance of the learned features for the target task.

\begin{figure}[!t]
    \centering
    \includegraphics[width=\linewidth]{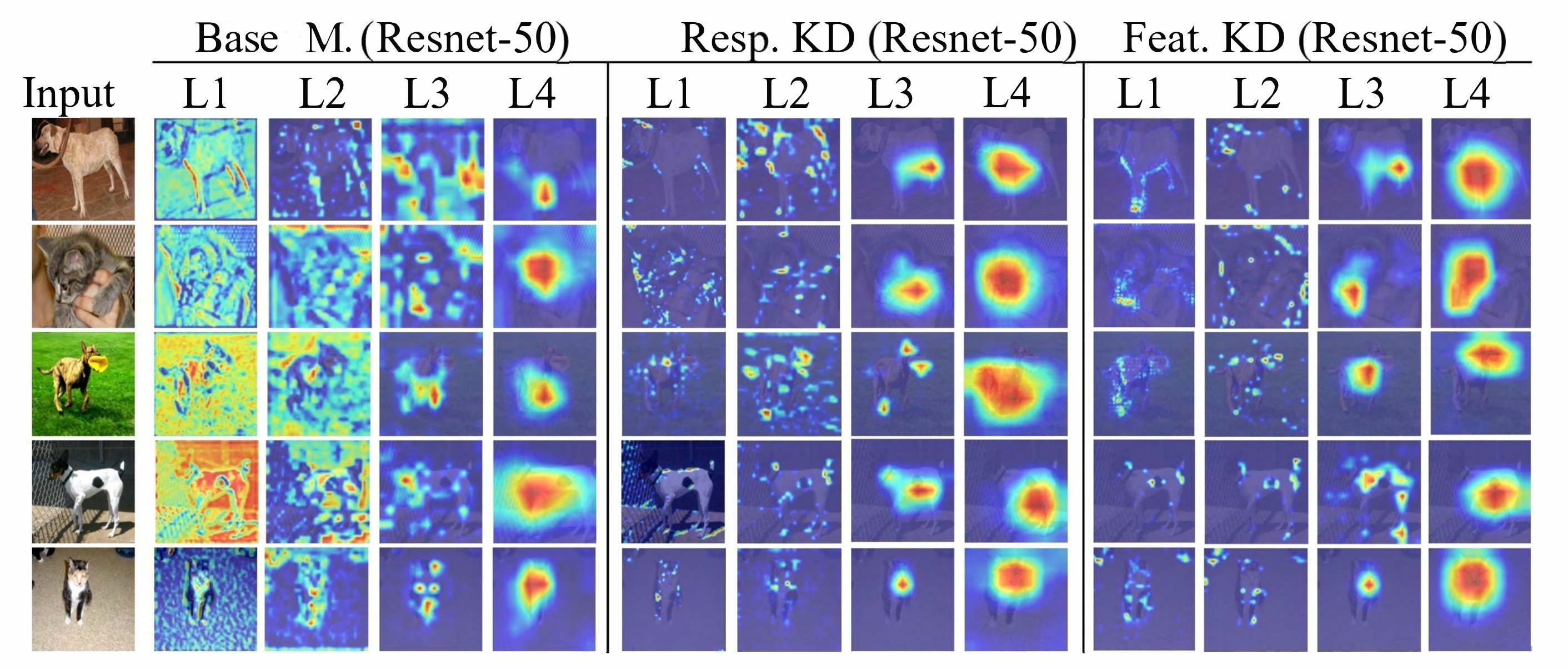}
    \caption[Grad-CAM in various KD Techniques]{Grad-CAM visualisation of the Base model and Student trained with various KD techniques.}
    \label{Fig:compareCam}
\end{figure}

Fig.~\ref {Fig:compareCam} shows the Grad-CAM visualisation at L1, L2, L3, and L4 of the last residual blocks in the four layers of the ResNet-50. The Base model relies on low-level features such as edges and spreads the attention over the entire image, including the background, in the first and intermediate layers. The saliency maps generated by Student models, however, localise more salient regions and focus on the object in all layers. This suggests that KD helps a model learn better features and improve its localisation ability by directing attention to more salient features earlier in the network.

\begin{figure}[!hb]
    \centering
    \includegraphics[width=0.98\linewidth]{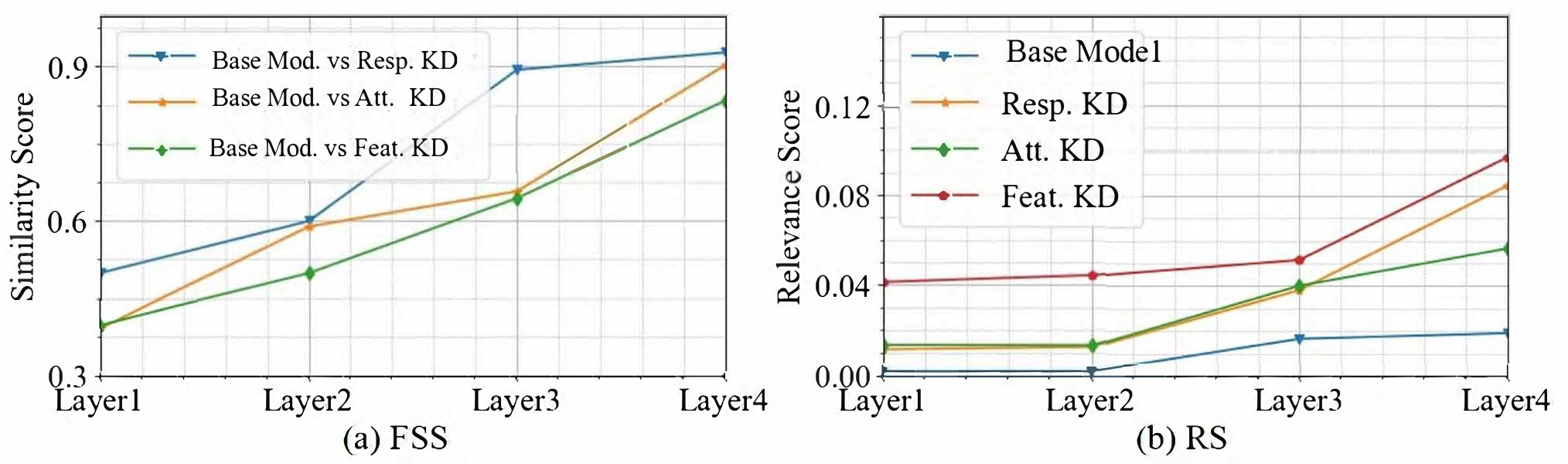}
    \caption[FSS and RS Scores]{(a) FSS and (b) RS between Student (ResNet-50) and Base model (ResNet-50), localised by Grad-CAM.}
    \label{Fig:compareCams}
\end{figure}

We then use \textit{FSS} and \textit{RS} metrics to quantify the attention pattern similarity and relevance of the features between the Student and Base models. Fig.~\ref {Fig:compareCams} (a) and (b) show the feature similarity of the attention patterns (\textit{FSS}) and their relevance score (\textit{RS}) between the Base model and Student across different layers, respectively. The \textit{FSS} is higher for the deeper layers than for the input and intermediate layers, indicating that the Student models either learn more salient features in the input layers or fail to learn more irrelevant features. However, the Grad-CAMs in Fig.~\ref{Fig:compareCam} show that the Student models have localised far better salient features than the Base model, especially in the input and intermediate layers. Therefore, the lower \textit{FSS} at input and intermediate layers suggests that the Students have learned more relevant features that the Base model has not learned yet. Moreover, the Student models achieve higher \textit{RS} than the Base model across all layers, implying that the models trained with KD have learned more relevant features with the guidance of the Teacher knowledge.

In general, Fig.~\ref {Fig:compareCam} and Fig.~\ref{Fig:compareCams} demonstrate that KD enables the Student models to encode more relevant features, which enhance the prediction accuracy and ability to generalise. 

\subsubsection{Visualising and quantifying distilled knowledge} 
\begin{figure}[!t] 
    \centering 
    \begin{subfigure}{\linewidth} 
    \includegraphics[width=\linewidth]{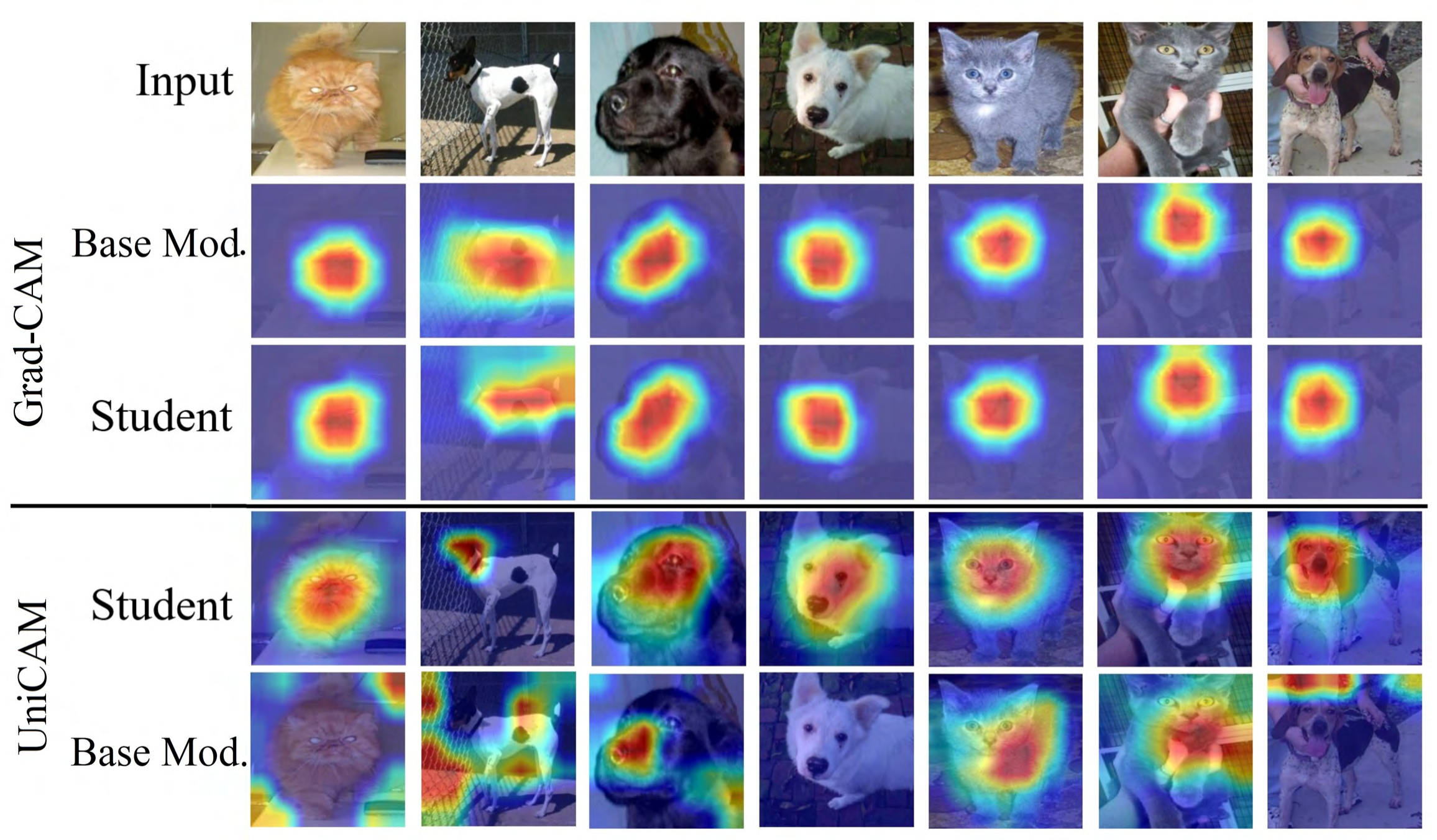} 
    \caption{Pet Images} 
    \vspace{0.5cm} 
    \end{subfigure} 
    \begin{subfigure}{\linewidth} 
    \includegraphics[width=\linewidth]{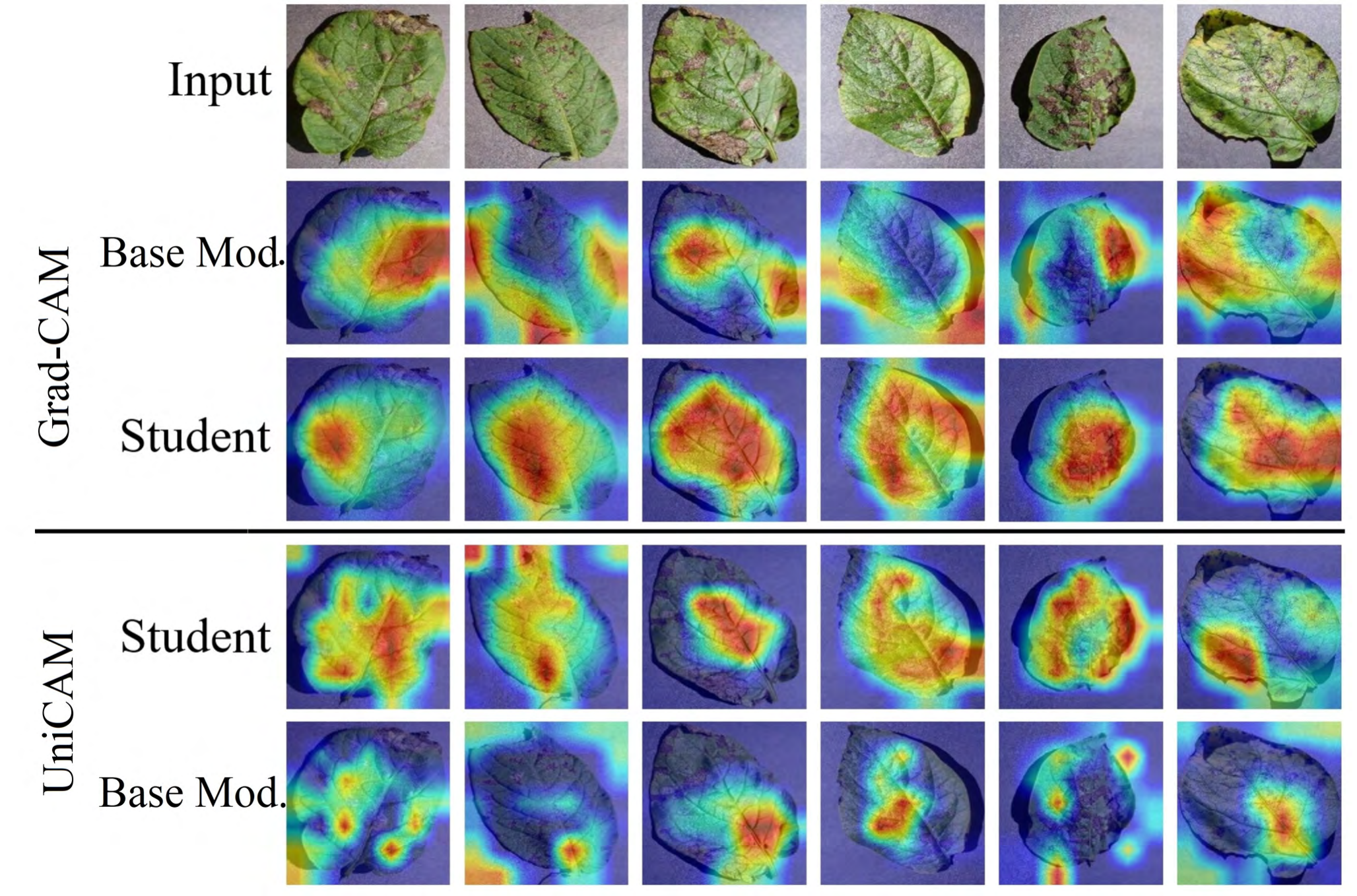} 
    \caption{Potato Early Blight} 
    \vspace{0.5cm} 
    \end{subfigure} 
    \caption{Sample visualisation of Distilled and residual features on PetImages and Plant disease dataset.}
    \label{Fig:visdiff} 
\end{figure}

Here, we use our proposed method, \textit{UniCAM}, to visualise the distilled and residual knowledge during KD. The saliency maps generated using \textit{UniCAM} show that KD is not a simple feature copying process from the Teacher to the Student but a guided training process where the Teacher’s knowledge assists the Student to learn existing or new features. This is illustrated in Fig.~\ref{Fig:visdiff}, where the distilled features mainly focus on the primary object, whereas the residual features localise regions in the background or seemingly less relevant parts of the object. In certain cases, \textit{UniCAM} does not highlight any part of the object in the Base model. This occurs because, after the removal of common features, the remaining features from the Base model are less significant or lack relevance to the target task. In the plant disease classification\footnote{More results for plant disease experiments are presented in \textbf{Sec. A} of the supplementary material}, distilled features accurately identify segments of leaves essential for disease classification, demonstrating that KD helps models learn more relevant features.

In addition, Table~\ref{tab:uniquefeatures} quantifies the relevance of the distilled, residual, and layer-specific features. Distilled and residual features are extracted from the areas localised by \textit{UniCAM}, while layer-specific features are extracted from regions localised using Grad-CAM. Distilled features indicate how relevant the knowledge learned from the Teacher is for the Student. In contrast, residual features represent features the Student deemed less useful for the task and thus ignored. Layer-specific features provide an overall view of features encoded at each layer, helping assess which model has better localisation and a higher \textit{RS}, indicating more relevant features. Models trained with various KD techniques show higher \textit{RS} than their equivalent Base models, with overhaul feature distillation achieving the best performance by transferring intermediate feature representations, allowing the Student to learn more fine-grained and diverse features.
\begin{table}[!t]
\caption[Quantifying the Relevance of Features]{Relevance of features (RS) learned by Student (ResNet-50) and Base model (ResNet-50).}
\label{tab:uniquefeatures}
\resizebox{\columnwidth}{!}{%
\begin{tabular}{lllcc|cccc}
\hline
\multirow{2}{*}{Dataset} &
  \multirow{2}{*}{KD-Technique} &
  \multirow{2}{*}{Layer\#} &
  \multicolumn{2}{c|}{Layer-Specific Features} &
  \multicolumn{2}{c}{Residual / Distilled Features} \\ \cline{4-7} 
  & & & Base Model & Student & Base Model & Student \\ \hline
\multirow{12}{*}{ASIRRA~\cite{elson2007asirra}} &
  \multirow{4}{*}{Response-based} &
  L1 &
  0.0092 &
  \textbf{0.0189} &
  \textbf{0.0024} &
  0.0017 \\
 &                                  & L2 & 0.0054          & \textbf{0.0130}   & 0.0001          & \textbf{0.0040}  \\
 &                                  & L3 & 0.0100             & \textbf{0.0365}  & 0.0007          & \textbf{0.008}  \\
 &                                  & L4 & 0.0141           & \textbf{0.0861}  & 0.0043          & \textbf{0.006}  \\ \cline{2-7} 
 & \multirow{4}{*}{Attention-based} & L1 & 0.0092           & \textbf{0.0107}  & \textbf{0.0049} & 0.0047          \\
 &                                  & L2 & 0.0054           & \textbf{0.0189}  & 0.0022          & \textbf{0.0035} \\
 &                                  & L3 & 0.0100             & \textbf{0.0431}  & 0.0045          & \textbf{0.0100}   \\
 &                                  & L4 & 0.0141           & \textbf{0.0583}  & 0.0082          & \textbf{0.0102} \\ \cline{2-7} 
 & \multirow{4}{*}{Feature-based}   & L1 & 0.0092           & \textbf{0.0465}  & 0.0063          & \textbf{0.0101} \\
 &                                  & L2 & 0.0054           & \textbf{0.0453}  & 0.0027          & \textbf{0.0048} \\
 &                                  & L3 & 0.0100             & \textbf{0.0570}   & 0.0036          & \textbf{0.0196} \\
 &                                  & L4 & 0.0141           & \textbf{0.0953}  & 0.0012          & \textbf{0.0258} \\ \hline
\multirow{12}{*}{CIFAR10~\cite{krizhevsky2009learning}} &
  \multirow{4}{*}{Response-based} &
  L1 &
  0.0063 &
  \textbf{0.0304} &
  0.0040 &
  \textbf{0.0155} \\
 &                                  & L2 & 0.0133           & \textbf{0.0378}  & 0.0090           & \textbf{0.0148} \\
 &                                  & L3 & 0.0282           & \textbf{0.0432}  & 0.0046          & \textbf{0.0113} \\
 &                                  & L4 & 0.0417           & \textbf{0.0585}  & 0.0090           & \textbf{0.0106} \\ \cline{2-7} 
 & \multirow{4}{*}{Attention-based} & L1 & 0.0063           & \textbf{0.0232}  & 0.0043          & \textbf{0.0136} \\
 &                                  & L2 & 0.0133           & \textbf{0.0280}   & 0.0099          & \textbf{0.0101} \\
 &                                  & L3 & \textbf{0.0282}  & 0.0256           & \textbf{0.0087} & 0.0063          \\
 &                                  & L4 & 0.0417           & \textbf{0.0437}  & 0.0017          & \textbf{0.0021} \\ \cline{2-7} 
 & \multirow{4}{*}{Feature-based}   & L1 & 0.0063            & \textbf{0.0311}  & 0.0028          & \textbf{0.0185} \\
 &                                  & L2 & 0.0133           & \textbf{0.0388}  & 0.0017          & \textbf{0.0153} \\
 &                                  & L3 & 0.0282           & \textbf{0.0457}  & 0.0070           & \textbf{0.0117} \\
 &                                  & L4 & 0.0417           & \textbf{0.0794}  & 0.0024          & \textbf{0.0150}  \\ \hline
\end{tabular}%
}
\end{table}

\subsubsection{Exploring the capacity gap impact} 
The Student's performance often declines when there is a large architecture (capacity) gap between the Teacher and the Student~\cite{stanton2021does,mirzadeh2020improved}. The drop in the Student's performance may stem from either its own challenges in learning relevant features or the overwhelming knowledge of the Teacher. To investigate this issue, we employ two distillation strategies in our experiments using ResNet-101 as the Teacher and ResNet-18 as the Student, which have a significant capacity disparity. In the first approach, we conduct direct distillation from ResNet-101 to ResNet-18. The second approach introduces an intermediate ``Teacher assistant"~\cite{Son2021ICCV} to help bridge the capacity gap between ResNet-101 and ResNet-18. We use \textit{UniCAM} and \textit{RS} to analyse the KD process in these settings, with a focus on how well the smaller model manages to learn relevant features.

\begin{figure}[!t]
    \centering
    \includegraphics[width=\linewidth]{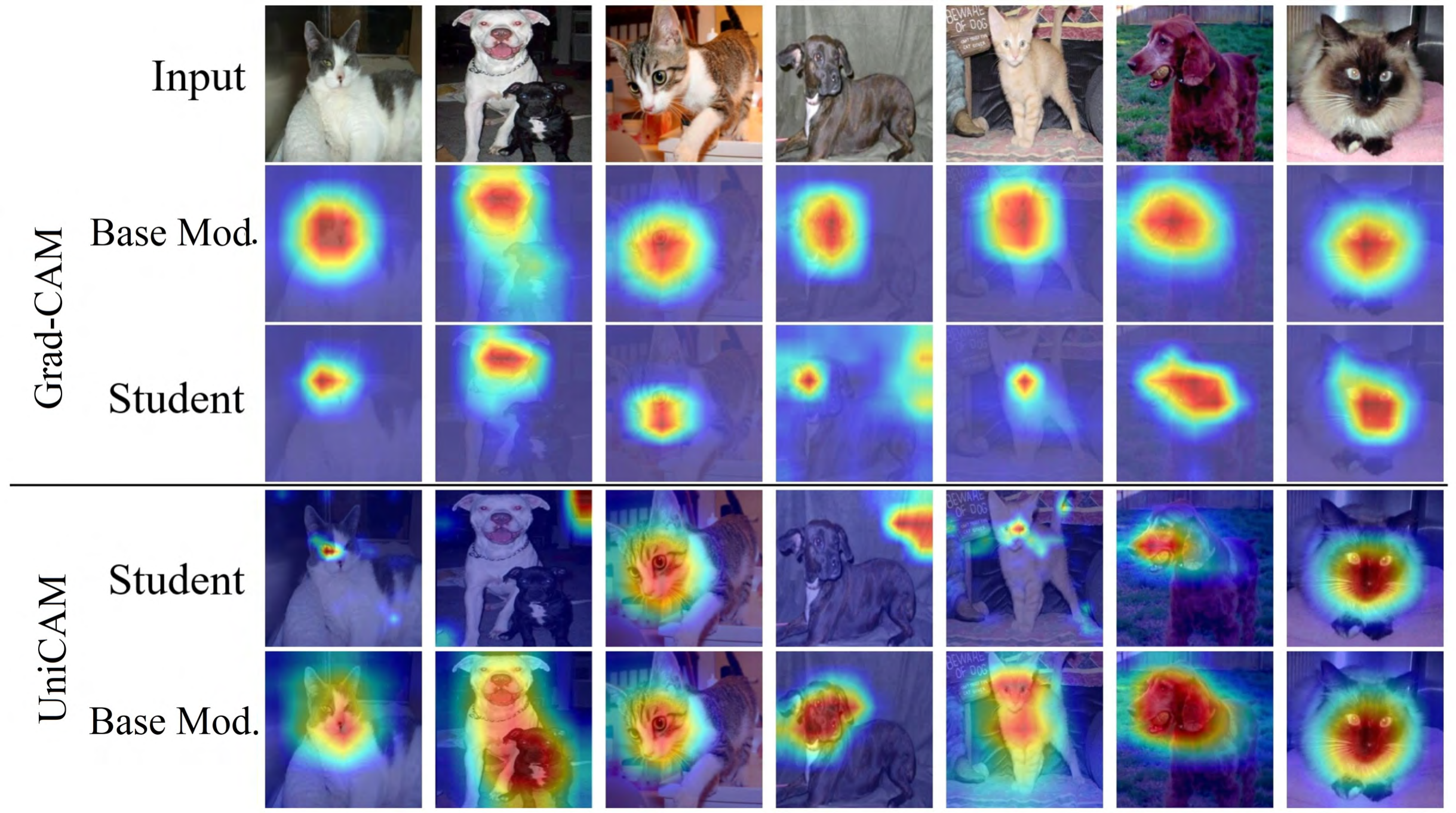}
    \caption{Relevant features learned by Student (ResNet-18) distilled from ResNet-101 compared to Base Model.}
    \label{Fig:supliment5}
\end{figure}

Using the proposed methods, we first examine the impact of a large capacity gap on the knowledge transfer between Teacher and Student. We use ResNet-101 as the Teacher and ResNet-18 as the Student and apply KD to train the Student model. Fig.~\ref{Fig:supliment5} demonstrates that, in this setting, the Base Model captures more relevant features than the Student model. This suggests that a large capacity gap impedes knowledge transfer, as the Student model cannot effectively learn from the complex Teacher's knowledge.

\begin{figure}[!ht]
    \centering
    \includegraphics[width=\linewidth]{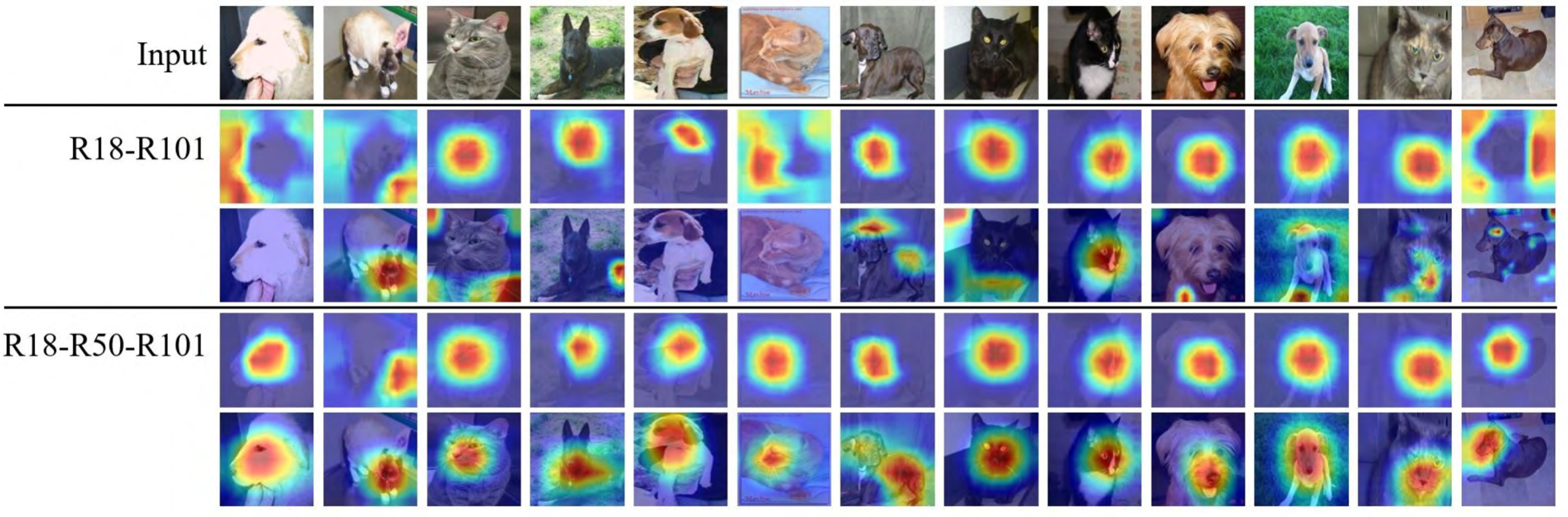}
    \caption[Comparing Student models distilled distilled from various setups]{Grad-CAM ($2^{nd}$ and $4^{th}$ rows) and \textit{UniCAM} ($3^{rd}$ and $5^{th}$ rows) visualisations.}
    \label{Fig:varrArch}
\end{figure}

To bridge the capacity gap, we use an intermediate Teacher assistant~\cite{mirzadeh2020improved} to enable a more effective and focused knowledge transfer from ResNet-101 to ResNet-18 via ResNet-50. Figure~\ref{Fig:varrArch} compares the saliency maps of the distilled features learned by two Students: ResNet-18 directly distilled from ResNet-101 (R18-R101) and ResNet-18 distilled from ResNet-101 through Teacher assistant ResNet-50 (R18-R50-R101). The saliency maps, visualised using \textit{UniCAM}, reveal that the Teacher assistant helps learn more relevant features that highlight the object parts. In contrast, R18-R101 learns some irrelevant features and misses the salient features for the $gt$ prediction. In fact, incorporating the Teacher assistant model facilitates the Student model in learning compatible knowledge from the complex Teacher and provides more appropriate supervision and feedback.

\begin{figure}[!t]
    \centering
    \includegraphics[width=\linewidth]{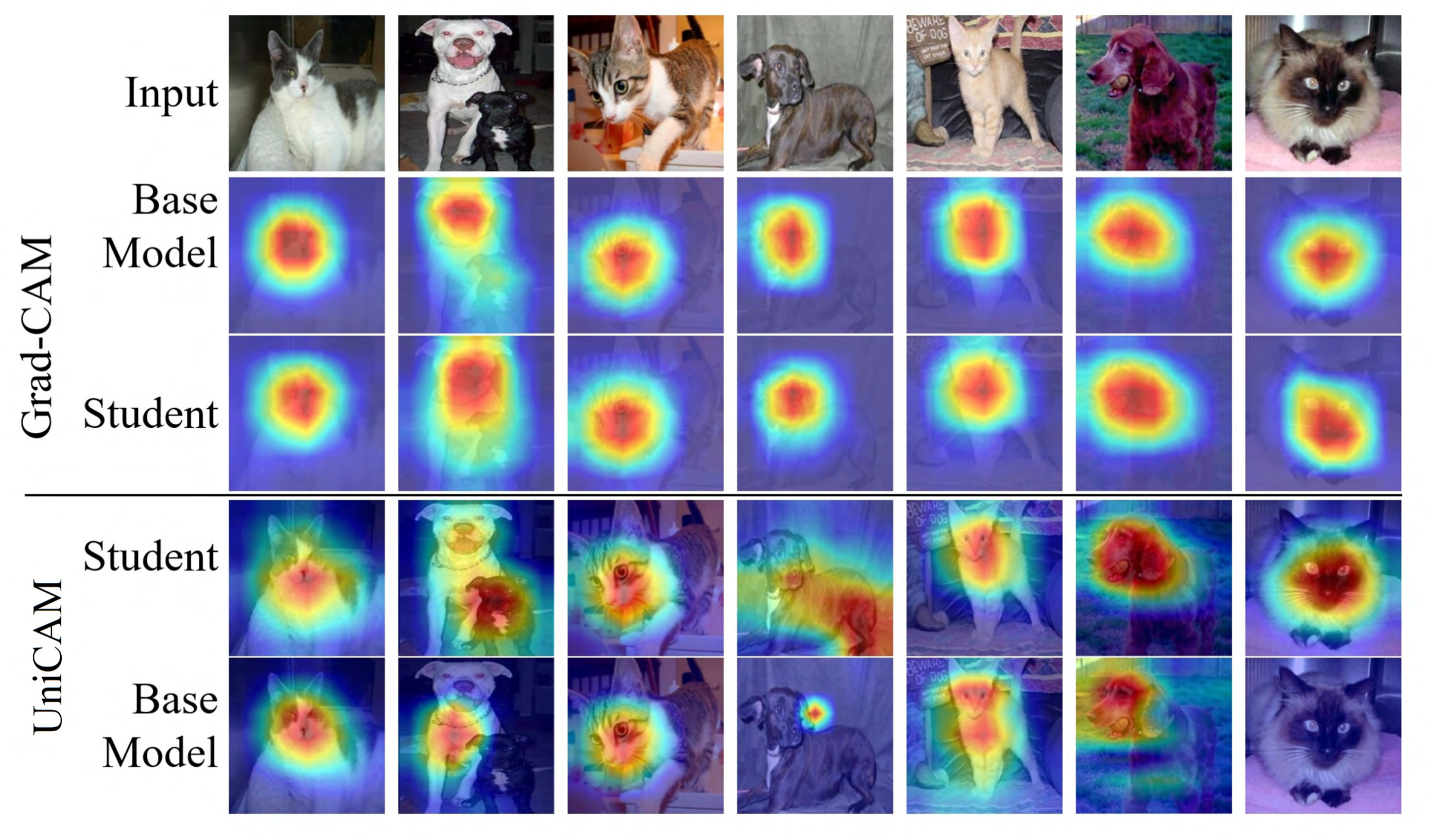}
    \caption{Comparison of distilled and residual features between Student (R18-R50-R101) and Base Model.}
    \label{Fig:supliment7}
\end{figure}

We compare the relevance of features learned by the Student trained with the Teacher assistant and its equivalent Base model. We used \textit{UniCAM} to generate the saliency maps of the distilled and residual features of each model. Fig.~\ref{Fig:supliment7} shows that the saliency maps of the distilled features are more focused on the salient regions of the input images, while the residual features are more dispersed.

\begin{table}[!htb]
\centering
\caption{RS of Students trained using Response-based KD with varying Teacher complexity level.}
\label{tab:compStd}
\resizebox{\columnwidth}{!}{%
\begin{tabular}{|c|c|c|c|c|c|c|}
\hline
\multirow{2}{*}{\textbf{Layer\#}} & \multicolumn{3}{c|}{\textbf{Layer-Specific Features}} & \multicolumn{3}{c|}{\textbf{Residual / Distilled Features}} \\ \cline{2-7} 
 & \textbf{Base model} & \textbf{R18-R101} & \textbf{R18-R50-R101} & \textbf{Base model} & \textbf{R18-R101} & \textbf{R18-R50-R101} \\ \hline
\textbf{Layer 1} & 0.0037 & 0.0022 & \textbf{0.0052} & 0.0014 & 0.0007 & \textbf{0.005} \\ \hline
\textbf{Layer 2} & 0.0039 & 0.0035 & \textbf{0.005} & 0.0016 & 0.0014 & \textbf{0.0031} \\ \hline
\textbf{Layer 3} & 0.0057 & 0.0045 & \textbf{0.0074} & 0.0012 & 0.0008 & \textbf{0.006} \\ \hline
\textbf{Layer 4} & 0.0063 & 0.0052 & \textbf{0.0082} & 0.0018 & 0.0011 & \textbf{0.0076} \\ \hline
\end{tabular}
}

\end{table}
Finally, Table~\ref{tab:compStd} quantifies the relevance of the features learned by the Base model and equivalent Student model at different layers. The model trained with the Teacher assistant has learned more relevant features compared to the model directly distilled from ResNet-101 and the Base model.

The empirical findings presented above indicate that the capacity gap between the Teacher and Student models influences the quality and efficiency of KD. We demonstrate the benefit of our methods to explain the Student model's behaviour, both when it succeeds and fails to learn relevant knowledge from the Teacher. Therefore, our visual explanation and metrics can help to select the optimal Teacher-Student pairs for improved performance. 

\section{Discussion and Future Works}
\label{sec:Chap5Discussion}
This paper presented novel techniques to explain and quantify the knowledge during KD. We proposed \textit{UniCAM}, a gradient-based visual explanation method to explain the distilled knowledge and residual features during KD. Our experimental results show that \textit{UniCAM} provides a clear and comprehensive visualisation of the features acquired or missed by the Student during KD. We also proposed two metrics: \textit{FSS} and \textit{RS} to quantify the similarity of the attention patterns and the relevance of the distilled knowledge and residual features. The proposed method has certain limitations. The experiments were exclusively conducted on classification tasks, which is one of the many potential applications of KD. In addition, we acknowledge the added computational cost introduced by the need to compute pairwise distances, gradients for feature localisation, and the proposed metrics. As part of future work, we aim to extend \textit{UniCAM} to more complex datasets and explore its applicability to tasks beyond classification to enhance the robustness and versatility of the proposed method.
\section*{Acknowledgement}
This research was supported by ``PID2022-138721NB-I00'' grant from the Spanish Ministry of Science, Research National Agency and FEDER (UE).
\clearpage
{\small
\bibliographystyle{ieee_fullname}
\bibliography{egbib}
}

\clearpage
\newpage
\appendix

\section{Additional Results: Explaining distilled and residual features on Plant Disease Dataset}
\setcounter {equation} {0}
\setcounter {figure} {0}
\renewcommand {\theequation} {A.\arabic {equation}}
\renewcommand {\thefigure} {A.\arabic {figure}}
In the experimental sections, we evaluated our proposed method to explain and quantify Knowledge Distillation (KD) on well-established datasets such as ASIRA and CIFAR10, showing its versatility and effectiveness in various scenarios. To further validate the generalisability of our method, especially for identifying distilled and residual features, we also applied it to the plant disease classification. This additional analysis confirmed the suitability of the proposed explainability technique in challenging datasets. In this section, we summarise the results, highlighting the ability of our method to detect salient features essential for diagnosing plant diseases and demonstrating its wide applicability to real-world problems.

\begin{figure}[!htb]
    \centering
    \includegraphics[width=0.98\linewidth]{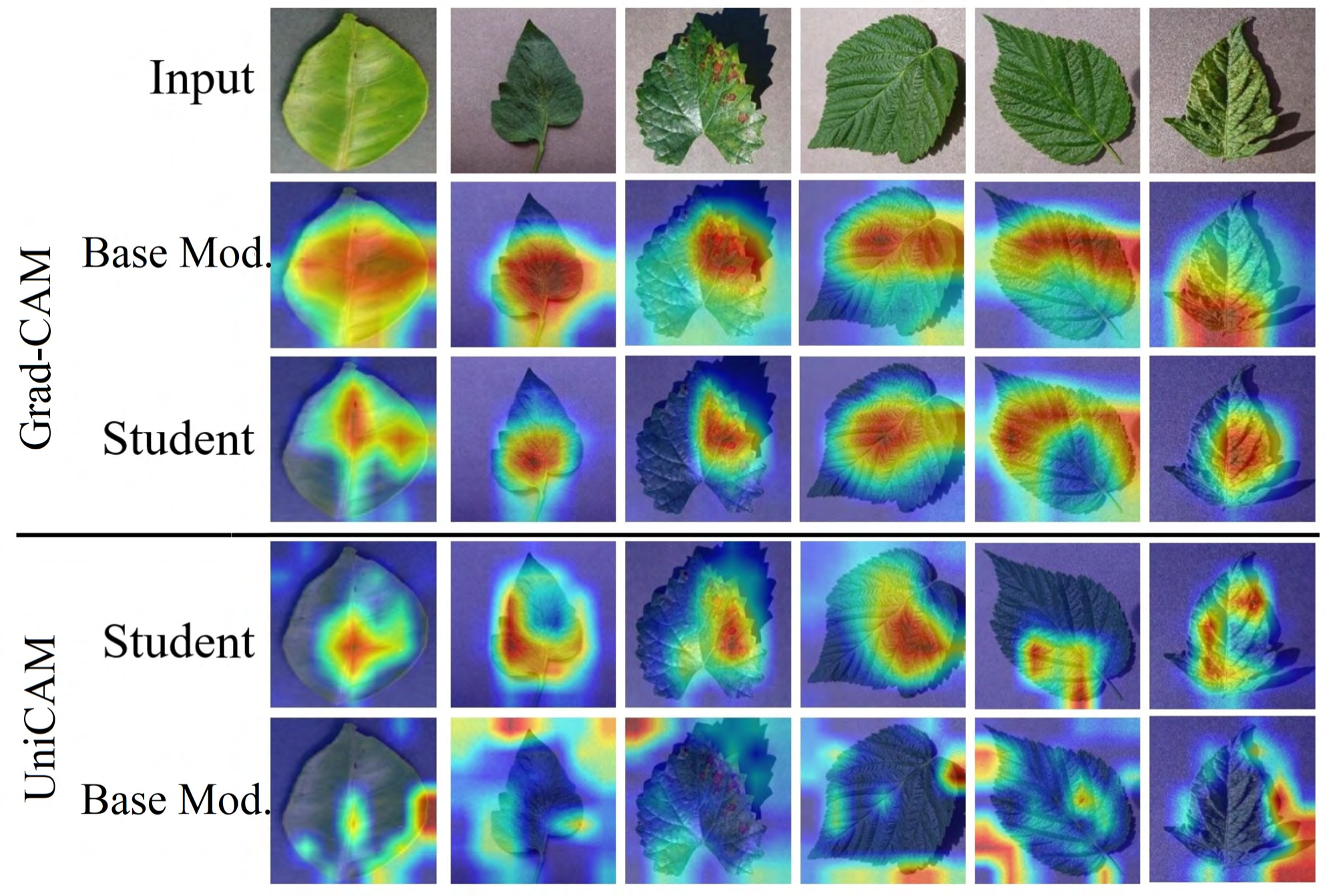}
    \caption{Sample visualisation of unique (distilled and residual) features in Plant disease classification.}
    \label{Fig:suplimentPlantDisease1}
\end{figure}

We generate additional results to visualise distilled and residual features for plant disease images. Fig.~\ref{Fig:suplimentPlantDisease1} shows that the Student model (ResNet-50) more accurately localises salient features compared to the Base model (ResNet-50). Specifically, the distilled features predominantly highlight regions relevant for accurate prediction, whereas the residual features tend to be distributed over areas irrelevant to the prediction. This implies that the Student model learns to ignore the features that are not useful for the prediction and focus on the salient parts. These saliency maps align with the findings presented in the main body of our work (in Fig. 4 and Fig. 6), explaining that the Student model consistently learns features of better relevance across different datasets compared to its equivalent Base model.

\begin{figure}[!htb]
    \centering
    \includegraphics[width=0.98\linewidth]{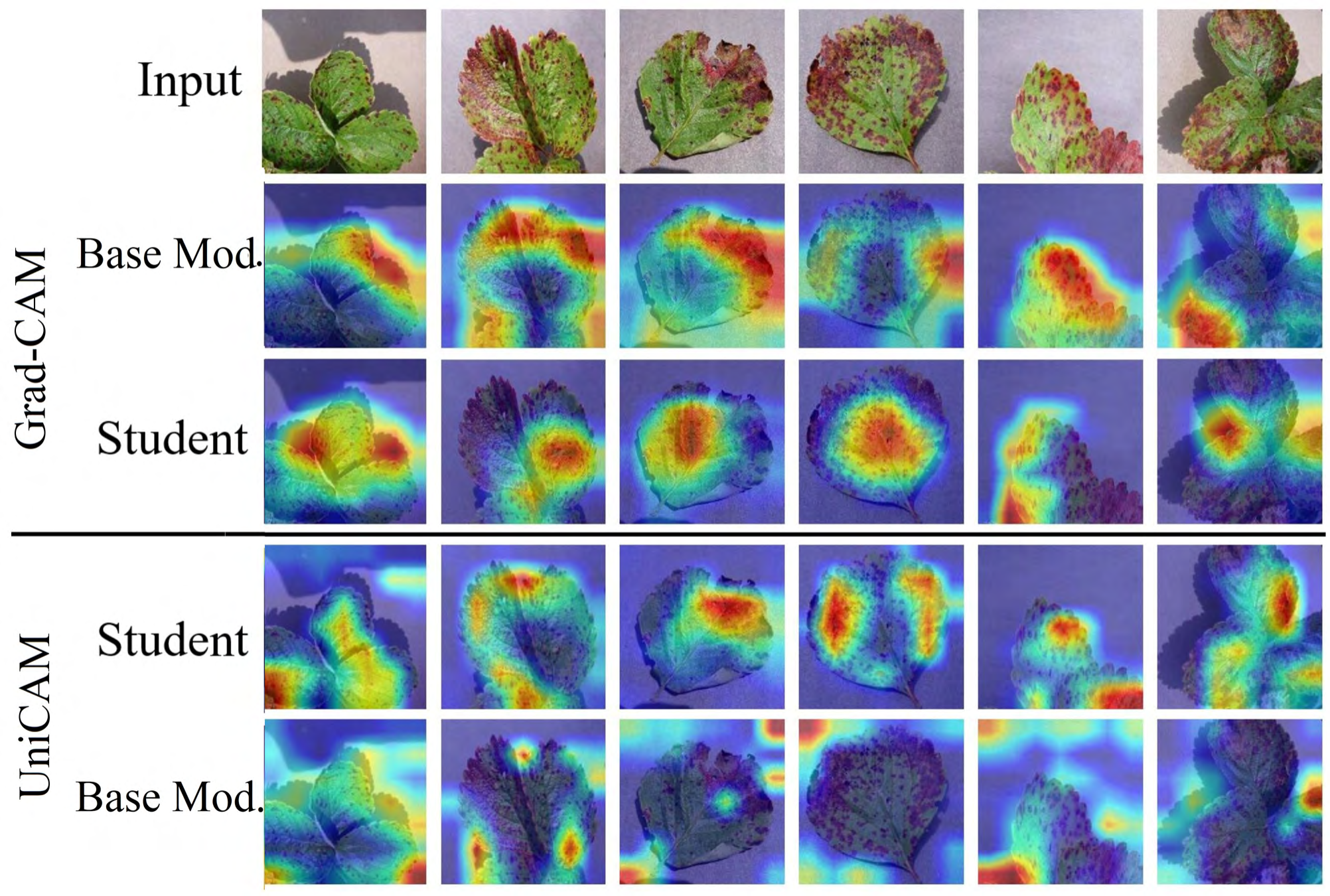} 
    \caption{Strawberry Leaf Scorch} 
    \vspace{0.5cm} 
    \caption{Sample visualisation of distilled and residual features on Strawberry Leaf Scorch plant disease classification.}
    \label{Fig:suplimentPlantDisease2} 
\end{figure}

Next, we analyse a specific case of the \textit{Strawberry Leaf Scorch} plant disease classification, and the Student shows an improved focus on the crucial signs of the disease on the leaves (Fig.~\ref{Fig:suplimentPlantDisease2}). The Student model can detect more relevant features for diagnosing Potato Early Blight and Strawberry Leaf Scorch.

\begin{figure}[!t] 
    \centering 
    \begin{subfigure}{\linewidth} 
    \includegraphics[width=\linewidth]{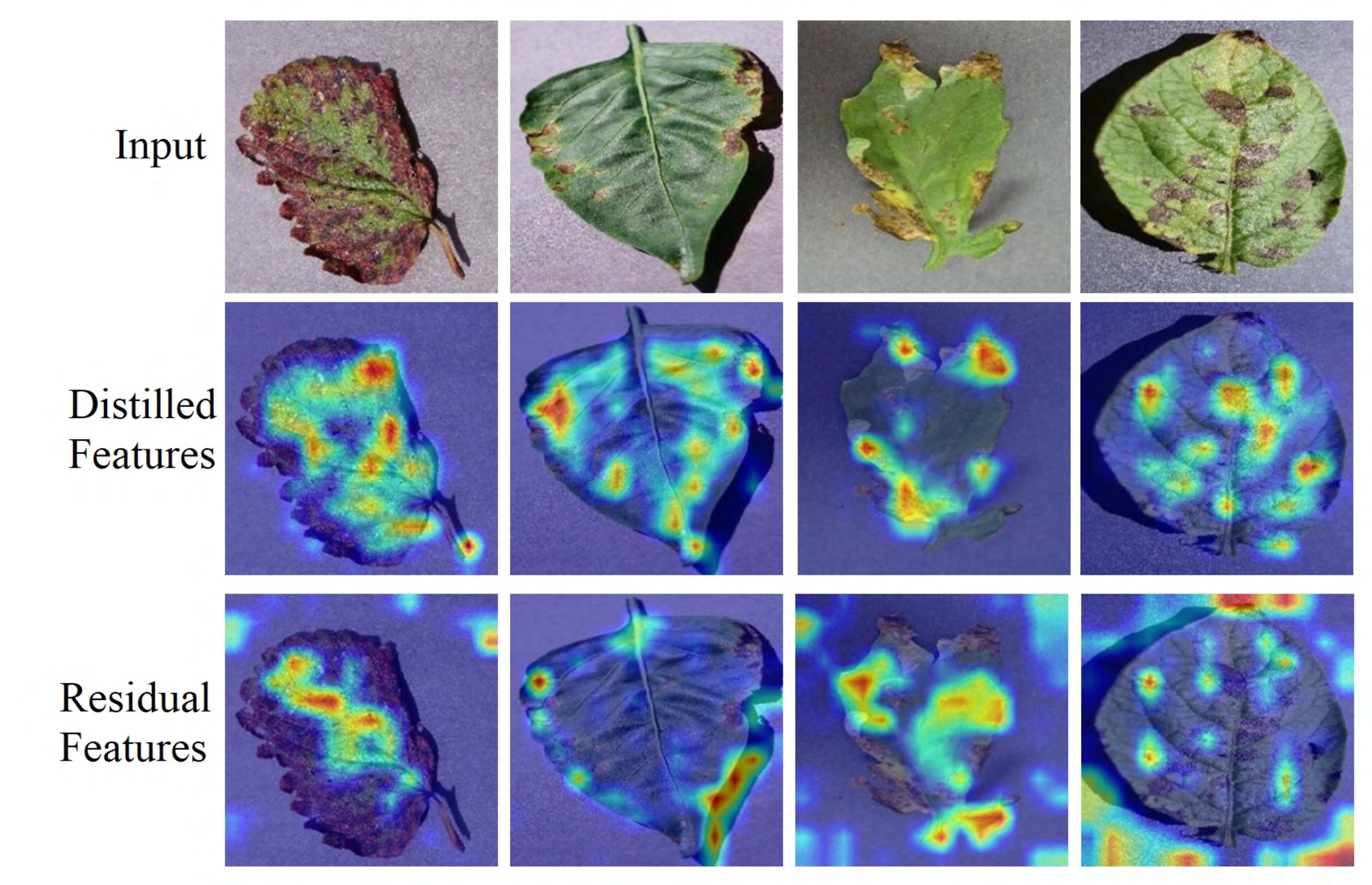} 
    \caption{Layer-3} 
    \vspace{0.5cm} 
    \end{subfigure} 
    \begin{subfigure}{\linewidth} 
    \includegraphics[width=\linewidth]{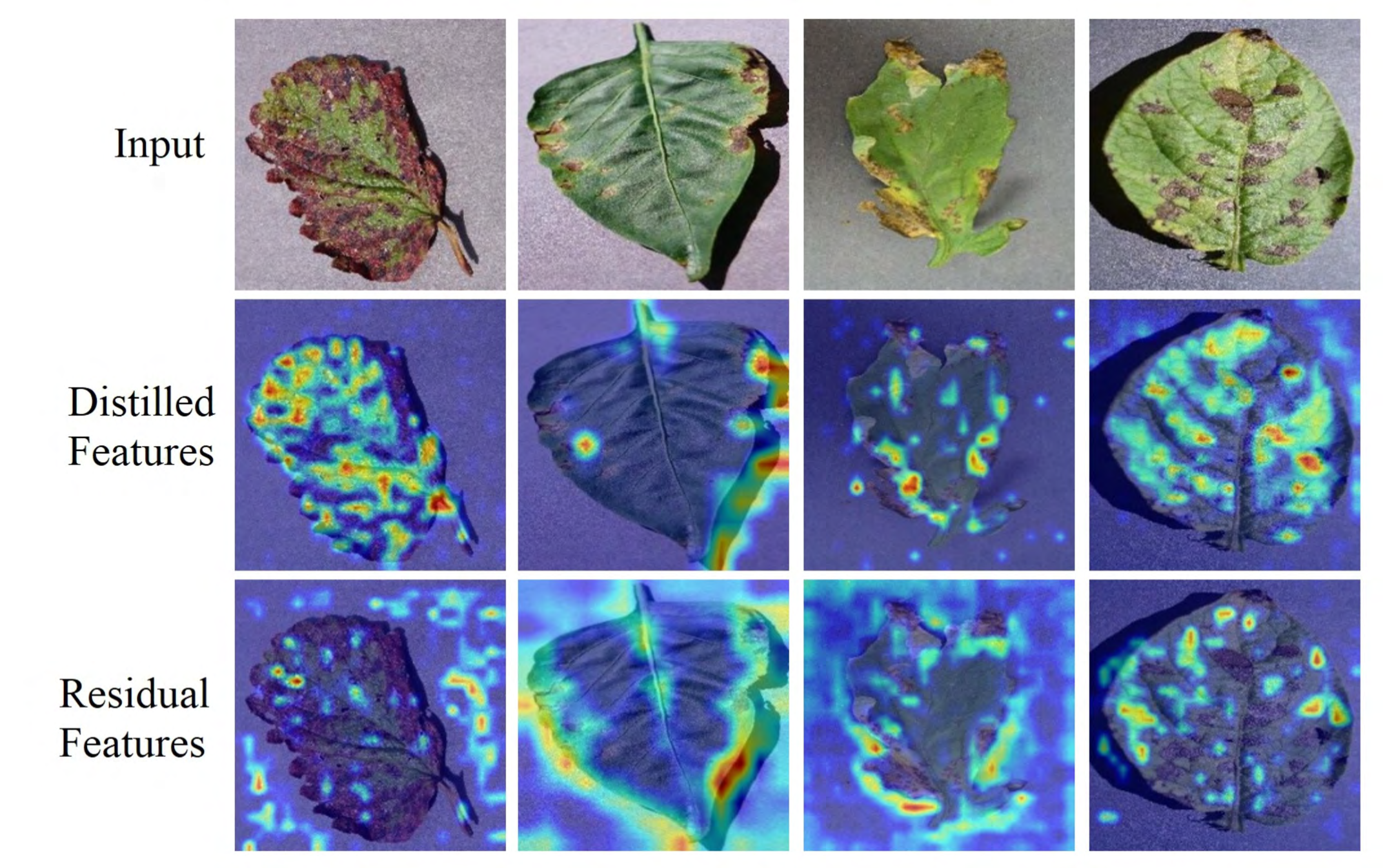} 
    \caption{Layer-2} 
    \vspace{0.5cm} 
    \end{subfigure} 
    \caption{Sample visualisation of Distilled and residual features on Plant disease classification from Layer-3 and Layer-2}
    \label{Fig:suplimentPlantDisease3} 
\end{figure}
In our further analysis, we visualised the distilled and residual features from Layer-3 and Lyer-2(Fig.\ref{Fig:suplimentPlantDisease3}). The distilled features mainly localise the diseased areas of the input image, even though they are challenging to locate. Meanwhile, the residual features highlighted the areas of the leaf image that had little impact on plant disease classification. To conclude, the proposed novel visual explanation and quantitative metrics help to explain and quantify the knowledge that the Student model learned and failed to learn from the Teacher model during Knowledge Distillation.

\section{Distance and Partial Distance Correlation}
\setcounter {equation} {0}
\setcounter {figure} {0}
\renewcommand {\theequation} {B.\arabic {equation}}
\renewcommand {\thefigure} {B.\arabic {figure}}
Here, we provide the detailed steps of distance and partial distance correlation following Szelkely et al.~\cite{szekely2014partial}. Distance correlation (dCor) measures linear and nonlinear associations or dependence between two random vectors. For an observed random sample $(x, y) = {(X_k, Y_k) : k = 1, \ldots , n}$ drawn from a distribution of random vectors, the empirical distance correlation $\ R_{n}^{2}(x,y)$ for $n$ samples is derived from the distance covariance of the samples. To compute the distance covariance between the samples, we first compute the $n$ by $n$ distance matrices $(a_{j, k})$ and $(b_{j, k})$ containing all pairwise distances:
\begin{equation} \label{eq:pairw1}
\begin{split}
&a_{j,k} = \left \|X_j - X_k \right \|, j,k = 1,2,\cdots, n, \\
&b_{j,k} = \left \|Y_j - Y_k \right \|, j,k = 1,2,\cdots, n
 \end{split}
\end{equation}
where $\|.\|$ represents the Euclidean norm. Taking all the doubly centred distances as:
\begin{equation} \label{eq:doubly1}
\begin{split}
&A_{j,k} := a_{j,k} - \overline{a}_{j\cdot } - \overline{a}_{\cdot k} + \overline{a}_{\cdot \cdot},\\
&B_{j,k} := b_{j,k} - \overline{b}_{j\cdot } - \overline{b}_{\cdot k} + \overline{b}_{\cdot \cdot},
 \end{split}
\end{equation}
where $\overline {a}_{j\cdot }$ is the $j^{th}$ row mean, $\overline {a}_{\cdot k}$ is the $k^{th}$ column mean, and $\overline {a}_{\cdot \cdot}$ is the grand mean of the distance matrix of the sample $x$ (the notation is similar for $b$ of sample $y$), then we compute the squared sample distance covariance $V(x,y)$ as the arithmetic average of the products $A_{j,k} B_{j,k}$:
\begin{equation}\label{eq:dcov1}
    V_n^2(x,y) = \frac{1}{n^2}\sum_{j,k = 1}^{n}A_{j,k}B_{j,k}.
\end{equation}
and the distance variance $V(x)$ and $V(y)$ as follows:
\begin{equation}\label{eq:dvar11}
    \text{dVar}_n^2(x) := V_n^2(x,x) = \frac{1}{n^2} \sum_{j,k = 1}^{n} A_{j,k}^2.
\end{equation}
\begin{equation}\label{eq:dvar12}
    \text{dVar}_n^2(y):= V_n^2(y,y) = \frac{1}{n^2}\sum_{j,k = 1}^{n}B_{j,k}^2.
\end{equation}

Distance variance is a measure of the complexity or diversity of a single random vector, while distance covariance is a measure of the dependence or similarity between two random vectors. Distance variance is a special case of distance covariance when the two random vectors are identical, meaning that they have the same information and knowledge. In our context, distance covariance and distance variance functions are used to measure the amount of knowledge transferred from Teacher to student models during knowledge distillation. They calculate the degree of dependence or similarity between the Teacher and student models and the degree of complexity or diversity within each model based on their outputs or features. The meaning of the result is as follows:
\begin{itemize}
    \item A high distance covariance between Teacher and student models means that they have a high degree of similarity or alignment in their information and knowledge, which implies a successful knowledge transfer.
    \item A low distance covariance between Teacher and student models means that they have a low degree of similarity or alignment in their information and knowledge, which implies an unsuccessful knowledge transfer or a potential overfitting or underfitting problem.
    \item A high distance variance for either the Teacher or student model means that it has a high degree of complexity or diversity in its information and knowledge, which implies a high capacity or expressiveness of the model.
    \item A low distance variance for either the Teacher or student model means that it has a low degree of complexity or diversity in its information and knowledge, which implies a low capacity or expressiveness of the model.
\end{itemize}

\begin{definition}
(Distance correlation)~\cite{szekely2014partial}. For an observed random sample $(x, y) = {(X_k, Y_k) : k = 1, \ldots , n}$ drawn from a distribution of random vectors $X$ in $\mathbb{R}^p$ and $Y$ in $\mathbb{R}^q$, the empirical distance correlation $\ R_{n}^{2}(x,y)$ for $n$ samples is defined as:
\end{definition}

\begin{equation}\label{eq:ddcor1}
R_{n}^{2}(x,y) = \begin{cases}
\frac{V_{n}^{2}(x,y)}{\sqrt{V_{n}^{2}(x,x)V_{n}^{2}(y,y)}} &,V_{n}^{2}(x,x)V_{n}^{2}(y,y) > 0\\
0 &, V_{n}^{2}(x,x)V_{n}^{2}(y,y) = 0 \end{cases}
\end{equation}
where $V_n^2(x,y)$, $V_n^2(x,x)$ and $V_n^2(y,y)$ are the squared sample distance covariance.  Distance correlation is zero when the random vectors are independent and one when they are dependent, indicating a strong correlation between each other.

Distance correlation (dCor) is a measure of the similarity of the information contained in two random variables. However, in some situations, we may want to measure the association between two random vectors after adjusting for a third random vector. This leads to the concept of partial distance correlation (pdCor), which is an extension of dCor proposed by Szekely et al.~\cite{szekely2014partial}. They introduced a Hilbert space where the squared distance covariance is an inner product and showed how to obtain \textit{U-centered} matrices $\tilde{A}$ from the distance matrices $(a_{j,k})$ such that their inner product is the distance covariance.

\begin{definition}
    ($U-centered$ matrix)~\cite{szekely2014partial}: Let $A = (a_{j,k})$ be a symmetric, real valued $n \times n$ matrix with zero diagonal, $n > 2$. The U-centred matrix of $A$  at the $(j, k)^{th}$ entry is defined by:
\begin{equation}
\begin{split}
    \tilde{A}_{j,k} = \begin{cases}
    a_{j,k}-\frac{1}{n-2}\sum_{l=1}^{n}a_{l,k}-\frac{1}{n-2}\sum_{i=1}^{n}a_{j,i} \\
    +\frac{1}{(n-1)(n-2)}\sum_{l,i=1}^{n}a_{l,i}, &  j\neq k,\\
    0, & j = k
    \end{cases}
\end{split}
\end{equation}
and the inner product between $\Tilde{A},\Tilde{B}$ is defined as $(\tilde{A}\cdot\tilde{B}) := \frac{1}{n(n-3)}\sum_{j\neq k}\tilde{A}_{j,k}\tilde{B}_{j,k},$
\end{definition}

\begin{definition}
(Partial distance correlation)~\cite{szekely2014partial}: Let $(x, y, z)$ be random variables observed from the joint distribution of $(X, Y, Z)$, then the partial distance correlation between $x$ and $y$ controlling for $z$ (assuming it as a confounding variable) is given by:
\begin{equation}\label{eq:pddcor1}
R^{*2}(x, y; z) := \begin{cases}
\frac{(P_z^{\perp}(x) \cdot P_z^{\perp}(y))}{\|P_z^{\perp}(x)\|\|P_z^{\perp}(y)  \|}&, \|P_z^{\perp}(x)\| \cdot \|P_z^{\perp}(y)  \|\neq 0,\\
0 & ,Otherwise
\end{cases}
\end{equation}
where, $P_z^{\perp}(x) = \tilde{A}-\frac{(\tilde{A}\cdot\tilde{C})}{(\tilde{C}\cdot\tilde{C})}\tilde{C}$, $P_z^{\perp}(y) = \tilde{B}-\frac{(\tilde{B}\cdot\tilde{C})}{(\tilde{C}\cdot\tilde{C})}\tilde{C}$ denotes the orthogonal projection of $\tilde{A}(x)$ and $\tilde{B}(y)$ onto $\tilde{C}(z)^\perp$ respectively, and $(P_z^{\perp}(x) \cdot P_z^{\perp}(y)) = \frac{1}{n(n-3)}\sum_{j\neq k}(P_z^{\perp}(x)_{j,k})P_z^{\perp}(y))_{j,k})$ is sample partial distance covariance.
\end{definition}

To summarise, we utilised distance correlation (dCor), a robust statistical method, to measure the degree of association between the Base model and the Student during Knowledge Distillation (KD). Furthermore, we used partial distance correlation (pdCor) to extract the distilled and residual features. This approach, coupled with gradient-based visual explainability techniques, helped to propose \textit{UniCAM}, which explains the KD process and provides a better understanding of the knowledge it acquired and overlooked during KD.

\subsection{Theoretical Basis of Feature Subtraction in UniCAM}
The subtraction operation in the formulation of \textit{UniCAM} is used to remove the shared feature representations between the Student and the Base model or vice versa, identifying features unique to each model. This approach is conceptually similar to orthogonal projection in linear algebra, where a vector is decomposed into components: one that lies along a reference direction and another orthogonal to it. In this case, consider the features $ x_s $ (Student features) and $ x_b $ (Base model features). The shared features between $ x_s $ and $ x_b $ are represented by their projection:
\begin{equation} 
\text{proj}_{x_b}(x_s) = \frac{\langle x_s, x_b \rangle}{\langle x_b, x_b \rangle} x_b,
\end{equation}
where $ \langle \cdot, \cdot \rangle $ represents the inner product, and this term quantifies the component of $ x_s $ aligned with $ x_b $. 

Now, $ x_s $ can be decomposed into two orthogonal components:
1. The component is aligned with $ x_b $ (shared features): $ \text{proj}_{x_b}(x_s) $.
2. The component orthogonal to $ x_b $ (unique features): $ x_s|unique = x_s - \text{proj}_{x_b}(x_s) $.

Thus, the subtraction is valid and justified because:
\[
x_s = \text{proj}_{x_b}(x_s) + (x_s - \text{proj}_{x_b}(x_s)),
\]
where $ \text{proj}_{x_b}(x_s) $ identifies the shared features, and $ x_s - \text{proj}_{x_b}(x_s) $ gives the unique features.

In \textit{UniCAM}, we work in the transformed space of pairwise distance matrices and the features are adjusted for mutual influence using a U-centered distance matrix, $P^{(s)}$ and $P^{(b)}$. This provides a robust mechanism to capture the relational structure of the features rather than their absolute values. This approach is invariant to shifts or rotations in the feature space, and the analysis focuses on the geometric relationships between features. The shared features are calculated as follows:
\begin{equation} 
\text{Shared}_{x_s} = \frac{\langle P^{(s)}, P^{(b)} \rangle}{\langle P^{(b)}, P^{(b)} \rangle} P^{(b)}.
\end{equation}

The unique features of the Student, after removing the shared features, are:
\begin{equation} 
x_{s|unique} = P^{(s)} - \text{Shared}_{x_s}.
\end{equation}

This subtraction extracts the component of $P^{(s)}$ orthogonal to $P^{(b)}$, preserving only the unique features of $x_s$ that do not exist in $x_b$. The operation is mathematically valid due to the properties of vector spaces, where such decomposition is meaningful in terms of orthogonal projections.   

\section{Steps on Feature Extraction}
\setcounter {equation} {0}
\setcounter {figure} {0}
\renewcommand {\theequation} {C.\arabic {equation}}
\renewcommand {\thefigure} {C.\arabic {figure}}
In this section, we explain the step-by-step feature extraction from the relevant regions (which was briefly introduced in Eq. 7, Section 3.2 in the main paper). 
Given a Base model or a Student, then we can extract the features as follows:
\begin{equation}
    \label{eq:sft} \hat{x} = f(I \odot \mathcal{H}) 
\end{equation}
where $I$ is the input image, $\mathcal{H}$ is the saliency map generated using \textit{UniCAM}, $\odot$ is the element-wise multiplication operator, and $f$ is a feature extraction function. We extract the features by applying perturbation technique~\cite{rong2022consistent} that modifies the input image $I$ by replacing each pixel $I_{ij}$ with the weighted average of its neighbouring pixels in the highlighted region as follows:
\begin{equation} 
\label{eq:mft1}{I}'_{ij} = \sum_{k,l} w_{kl} I_{kl} 
\end{equation}
where $w_{kl}$ is a weight that depends on the relevance of pixel $I_{kl}$ for prediction. The relevance of each pixel is determined by the saliency map generated using \textit{UniCAM}, which produces a heatmap $\mathcal{H}$ that assigns a value to each pixel based on its contribution to the relevant features learned by one model. The higher the value, the more relevant the pixel is. The weight $w_{kl}$ is proportional to $\mathcal{H}$, such that: 
\begin{equation}
\label{eq:weight}
w_{kl} = \frac{\mathcal{H}_{kl}}{\sum_{k,l} \mathcal{H}_{kl}} 
\end{equation}

Therefore, we can write Eq.~\ref{eq:mft1} as:
\begin{equation} \label{eq:rmft1}{I}'_{ij} = \sum_{k,l} \frac{\mathcal{H}_{kl}}{\sum_{k,l} \mathcal{H}_{kl}} I_{kl}. \end{equation}

We can simplify this equation by using element-wise multiplication and division operators and rewrite Eq.~\ref{eq:rmft1} as follows:
\begin{equation}
\label{eq:elementwise}
{I}' = \frac{I \odot \mathcal{H}}{\sum \mathcal{H}},
\end{equation}
where $\sum \mathcal{H}$ is a scalar that represents the sum of all elements in $\mathcal{H}$. This equation shows how we obtain a modified image ${I}'$ that contains only the features of interest for each model. To extract these features into a vector representation, we apply a feature extraction function $f$ to ${I}'$:
\begin{equation}
\label{eq:featext1}
\hat{x} = f({I}').
\end{equation}

We substitute Eq.~\ref{eq:elementwise} into Eq.~\ref{eq:featext1} to obtain:
\begin{equation}
\label{eq:modfeatext1}
\hat{x} = f(\frac{I \odot \mathcal{H}}{\sum \mathcal{H}}). 
\end{equation}

Since $\sum \mathcal{H}$ is a scalar, we can ignore it for feature extraction purposes, as it does not affect the relative values of the pixels. Therefore, we simplify Eq.~\ref{eq:modfeatext1} to:
\begin{equation}
\label{eq:ftt1} \hat{x} = f(I \odot \mathcal{H}). 
\end{equation}
This is the step-by-step formulation to extract features from the regions identified as relevant using \textit{UniCAM} or other gradient-based visual explainability.

\begin{figure}[!ht]
    \centering
    \includegraphics[width=0.98\linewidth]{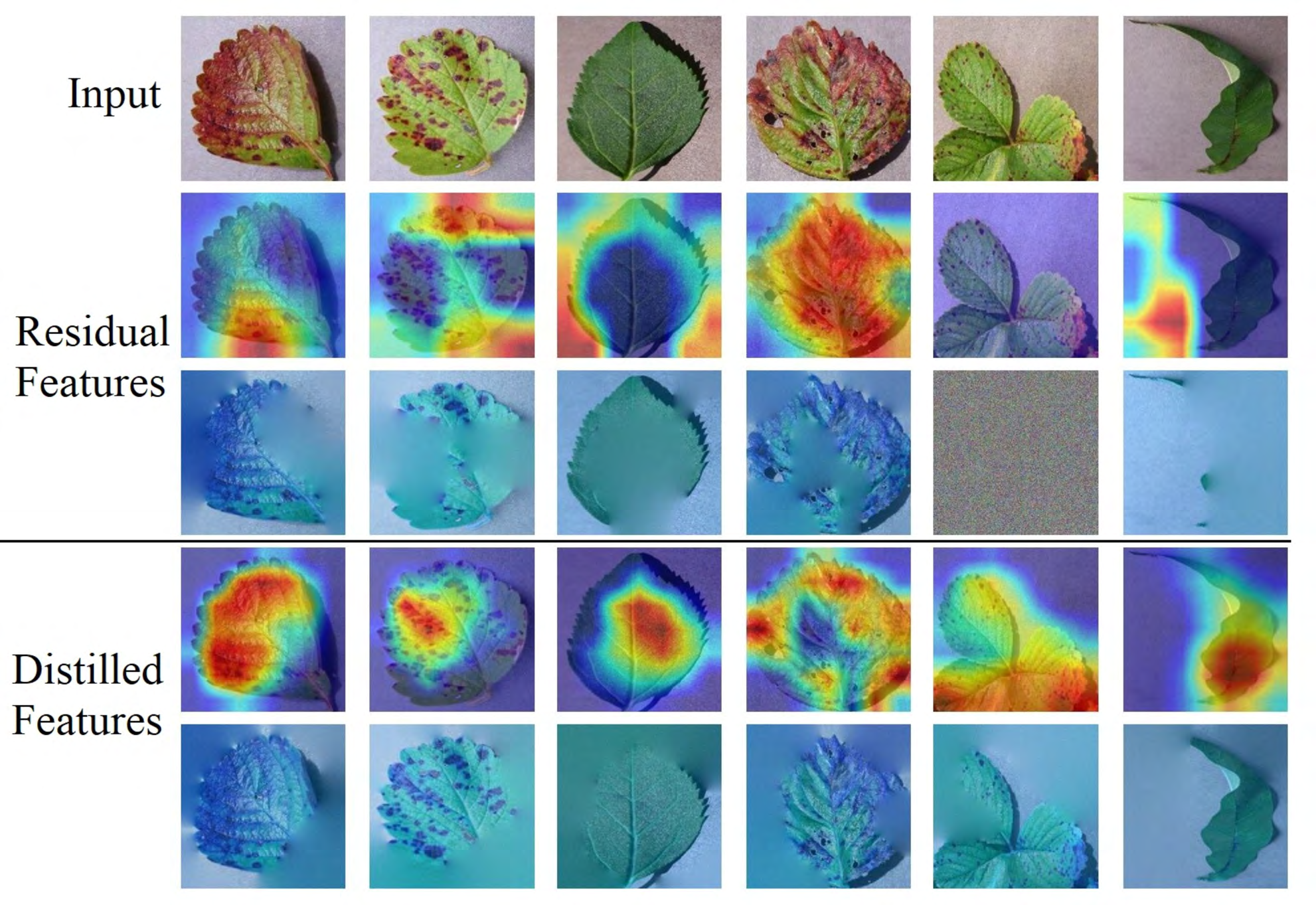}
    \caption{The residual and distilled features.}
    \label{Fig:suplimentfeatext}
\end{figure}

Fig.~\ref{Fig:suplimentfeatext} illustrates the results of the perturbed images $I' = I \odot \mathcal{H}$ for plant disease classification. First, we use \textit{UniCAM} to extract and explain the distilled and residual features from the original images. The distilled features focus on the diseased areas of the leaves, which are crucial for diagnosis, while the residual features are spread across the background or non-essential parts of the leaves. Next, we apply a perturbation technique proposed by Rong et al.~\cite{rong2022consistent} to modify the images based on pixel relevance. We then feed the modified images to the Student or Base model and obtain the corresponding feature vectors, $\hat{x}_s$ or $\hat{x}_b$. Finally, we use \textit{FSS} and \textit{RS} to measure the feature similarity and relevance of the feature vectors. The heatmap intensity for distilled features indicates a higher contribution to the classification decision, resulting in more distinct and less perturbed images of the critical areas for diagnosis after perturbation.

\setcounter {equation} {0}
\setcounter {figure} {0}
\renewcommand {\theequation} {D.\arabic {equation}}
\renewcommand {\thefigure} {D.\arabic {figure}}
\section{Experimental Details}
\subsection{Training Setup}
For our experiments, we employed three widely-used KD techniques: Response-based KD, Overhaul feature-based KD, and Attention-based KD. The training of the Student model followed an offline KD setup, where the Teacher model was pre-trained before being used to guide the Student. For the Teacher model, we used the Cross Entropy (CE) loss to optimise its performance based solely on the training data.

The Student model's training incorporated both the CE loss for the training data and the additional loss function specific to the KD technique applied, as recommended in their respective literature. For Response-based KD, the distillation loss minimised the divergence between the Teacher and Student outputs. Overhaul feature-based KD introduced intermediate feature-level supervision, and Attention-based KD utilised attention maps from the Teacher to align the Student's attention patterns.

In cases where the Student and Teacher shared the same architecture, we used the Teacher model as the Base model for comparison. When the Teacher and Student had different architectures, the Base model was trained using the same experimental settings as the Student, except without the Teacher's guidance, and optimised using only the CE loss.

\subsection{Selecting Layers}
We mainly employ ResNet family models, ResNet-18, ResNet-50, and ResNet-101, as the main convolutional neural network architectures for our implementation. We generate Grad-CAM and \textit{UniCAM} outputs from the last residual blocks in each of the four layers of the ResNet models. We denote these blocks as L1, L2, L3 and L4 and the details of these blocks are as follows:

\textbf{ ResNet-18:} This network has four layers with 2 residual blocks each. Each residual block has two convolutional layers, batch normalisation and ReLU activation. The first convolutional layer has a kernel size of $3 \times 3$ and preserves the number of channels. The second convolutional layer has a kernel size of $3 \times 3$ and also preserves the number of channels. The skip connection may have a convolutional layer to match the dimensions of the input and output. The last blocks in each layer have 64, 128, 256 and 512 output channels, respectively. Hence, each layer $L_i$ represents the following:
 \begin{itemize}
 \item \textbf{L1}: This block is the second and last block in the first layer. It has two convolutional layers with 64 output channels each.
\item \textbf{L2}: This block is the second and last block in the second layer. It has two convolutional layers with 128 output channels each.
\item \textbf{L3}: This block is the second and last block in the third layer. It has two convolutional layers with 256 output channels each.
\item \textbf{L4}: This block is the second and last block in the fourth layer. It has two convolutional layers with 512 output channels each.
\end{itemize}

\textbf{ResNet-50:} This network has four layers with 3, 4, 6 and 3 residual blocks, respectively. Each residual block has three convolutional layers with batch normalisation and ReLU activation. The first convolutional layer has a kernel size of $1 \times 1$ and reduces the number of channels by a factor of 4. The second convolutional layer has a kernel size of $3 \times 3$ and preserves the number of channels. The third convolutional layer has a kernel size of $1 \times 1$ and increases the number of channels by a factor of 4. The skip connection may also have a convolutional layer to match the dimensions of the input and output. The last blocks in each layer have 256, 512, 1024 and 2048 output channels, respectively.  Hence, each layer $L_i$ represents the following:
\begin{itemize} 
    \item \textbf{L1}: This block is the third and last block in the first layer. It has three convolutional layers with 64, 64 and 256 output channels, respectively. The skip connection does not have a convolutional layer.
    \item \textbf{L2}: This block is the fourth and last block in the second layer. It has three convolutional layers with 128, 128 and 512 output channels, respectively.
    \item \textbf{L3}: This block is the sixth and last block in the third layer. It has three convolutional layers with 256, 256 and 1024 output channels, respectively.
    \item \textbf{L4}: This block is the third and last block in the fourth layer. It has three convolutional layers with 512, 512 and 2048 output channels, respectively.
\end{itemize}

\textbf{ResNet-101:} This network has four layers with 3, 4, 23 and 3 residual blocks, respectively. Each residual block has three convolutional layers with batch normalisation and ReLU activation. The first convolutional layer has a kernel size of $1 \times 1$ and reduces the number of channels by a factor of 4. The second convolutional layer has a kernel size of $3 \times 3$ and preserves the number of channels. The third convolutional layer has a kernel size of $1 \times 1$ and increases the number of channels by a factor of 4. The skip connection may have a convolutional layer to match the dimensions of the input and output. The last blocks in each layer have 256, 512, 1024 and 2048 output channels, respectively. Hence, each layer $L_i$ represents the following:
 \begin{itemize}
\item \textbf{L1}: This block is the third and last block in the first layer. It has three convolutional layers with 64, 64 and 256 output channels, respectively.
\item \textbf{L2}: This block is the fourth and last block in the second layer. It has three convolutional layers with 128, 128 and 512 output channels, respectively. 
\item \textbf{L3}: This block is the twenty-third and last block in the third layer. It has three convolutional layers with 256, 256 and 1024 output channels, respectively.
\item \textbf{L4}: This block is the third and last block in the fourth layer. It has three convolutional layers with 512, 512 and 2048 output channels, respectively. 
\end{itemize}

\end{document}